%% file: main.tex
\documentclass[10pt,twocolumn,letterpaper]{article}

\usepackage[pagenumbers]{cvpr} 

\usepackage{graphicx}
\usepackage{amsmath}
\usepackage{amssymb}
\usepackage{booktabs}
\usepackage[table]{xcolor}
\usepackage{tabulary}
\usepackage{cuted}

\usepackage{xcolor}

\usepackage[pagebackref,breaklinks,colorlinks]{hyperref}
\usepackage{multirow}

\usepackage[capitalize]{cleveref}
\crefname{section}{Sec.}{Secs.}
\Crefname{section}{Section}{Sections}
\Crefname{table}{Table}{Tables}
\crefname{table}{Tab.}{Tabs.}

\def\cvprPaperID{4833} 
\def\confName{CVPR}
\def\confYear{2023}

\newcommand{\deemph}[1]{{\footnotesize{\color{black!40}#1}}}
\newcommand{\dataname}{PACO}
\newcommand{\datalvis}{PACO-LVIS}
\newcommand{\dataego}{PACO-EGO4D}

\newcommand{\apobj}{{$AP^{obj}$ }}
\newcommand{\apopart}{{$AP^{opart}$ }}

\newcommand{\attrexp}{\ref{sec:attribute_exp}}
\newcommand{\suppref}{appendix}
\newcommand\blfootnote[1]{
  \begingroup
  \renewcommand\thefootnote{}\footnote{#1}
  \addtocounter{footnote}{-1}
  \endgroup
}

\begin{document}

\title{\dataname{}: Parts and Attributes of Common Objects}

\author{Vignesh Ramanathan$^{*1}$ \quad Anmol Kalia$^{*1}$ \quad Vladan Petrovic$^{*1}$ \quad Yi Wen$^1$ \quad Baixue Zheng$^1$ \\
Baishan Guo$^1$ \quad Rui Wang$^1$ \quad Aaron Marquez$^1$ \quad Rama Kovvuri$^1$ \quad Abhishek Kadian$^1$ \\
Amir Mousavi$^{2\dagger}$ \quad Yiwen Song$^1$ \quad Abhimanyu Dubey$^1$ \quad Dhruv Mahajan$^1$ \\
$^1$Meta AI \quad $^2$Simon Fraser University \\
{\tt\small vigneshr@meta.com}
}
\maketitle
\blfootnote{$^*$ Equal contribution \enspace $^\dagger$ Work done during internship at Meta AI}

\begin{strip}
\vspace*{-12mm}
\centering\noindent
\includegraphics[width=1.0\textwidth]{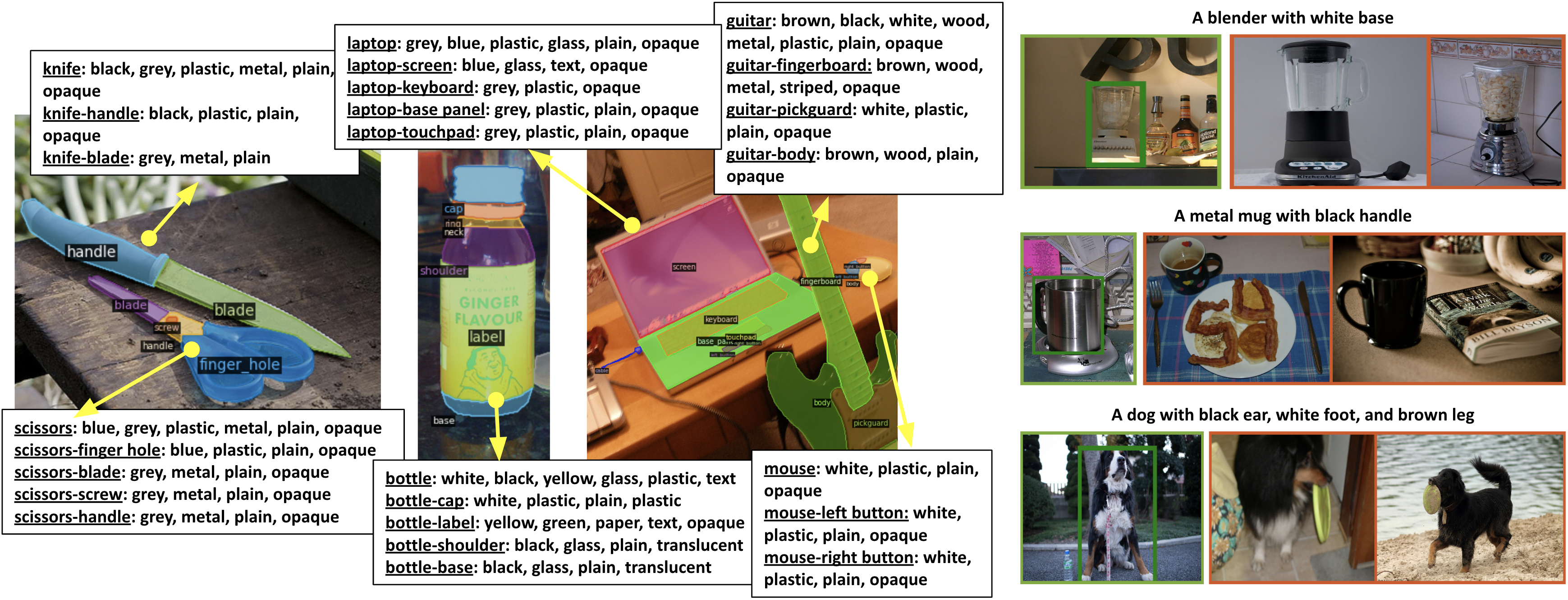}
\captionof{figure}{(left) \dataname{} includes objects with object masks, object attributes, part masks, and part attributes. (right) Object instance queries composed of object and part attributes are shown with corresponding positive images in green and negative images in red.}
\label{fig:pull_figure}
\end{strip}

\begin{abstract}
\vspace{-3mm}
Object models are gradually progressing from predicting just category labels to providing detailed descriptions of object instances. This motivates the need for large datasets which go beyond traditional object masks and provide richer annotations such as part masks and attributes. Hence, we introduce \dataname{}: Parts and Attributes of Common Objects. It spans $75$ object categories, $456$ object-part categories and $55$ attributes across image (LVIS) and video (Ego4D) datasets. We provide $641K$ part masks annotated across $260K$ object boxes, with roughly half of them exhaustively annotated with attributes as well. We design evaluation metrics and provide benchmark results for three tasks on the dataset: part mask segmentation, object and part attribute prediction and zero-shot instance detection. Dataset, models, and code are open-sourced at \href{https://github.com/facebookresearch/paco}{https://github.com/facebookresearch/paco}.
\vspace{-4mm}

\end{abstract}

\input{intro}

\input{related}
\input{dataset_construction}

\input{dataset_stats}
\input{evaluation_benchmark}

\input{experiments}

\section{Conclusion}
We introduced \dataname{}, a dataset designed to enable research towards joint detection of objects, parts and attributes of common objects. It provides part masks and attributes for $75$ common object categories spanning both image and video datasets. We introduce three benchmark tasks which showcase unique challenges in the dataset. Unlike object detection, these tasks require algorithms to cope better with smaller masks belonging to parts and have features that are not invariant to instance-level attributes. For all tasks, we provide results from extensions of existing detection models to help calibrate future research on the dataset.


{\small
\bibliographystyle{ieee_fullname}
\bibliography{main}
}
\input{appendix}

\end{document}

%% file: intro.tex
\section{Introduction}
\label{sec:intro}

Today, tasks requiring fine-grained understanding of objects like open vocabulary detection~\cite{kamath2021mdetr,feng2022promptdet,gupta2022ow,zhou2022detecting}, GQA~\cite{hudson2019gqa}, and referring expressions~\cite{kazemzadeh2014referitgame,mao2016generation,Chen_2020_CVPR} are gaining importance besides traditional object detection. Representing objects through category labels is no longer sufficient. A complete object description requires more fine-grained properties like object parts and their attributes, as shown by the queries in Fig.~\ref{fig:pull_figure}. 


Currently, there are no large benchmark datasets for common objects with joint annotation of part masks, object attributes and part attributes (Fig.~\ref{fig:pull_figure}). Such datasets are found only in specific domains like clothing ~\cite{zheng2018modanet,jia2020fashionpedia}, birds~\cite{wah2011caltech} and pedestrian description~\cite{li2016richly}. Current datasets with part masks for common objects~\cite{chen2014detect,zhou2019semantic,he2021partimagenet} are limited in number of object instances with parts ($59K$ for ADE20K~\cite{chen2014detect} Tab.~\ref{tab:dataset_stats}). On the attributes side, there exists large-scale datasets like Visual Genome~\cite{krishna2017visual}, VAW~\cite{pham2021learning} and COCO-attributes~\cite{patterson2016coco} that provide object-level attributes. However, none have part-level attribute annotations.

In this work, we enable research for the joint task of object detection, part segmentation, and attribute recognition, by designing a new dataset: \dataname{}. With video object description becoming more widely studied as well~\cite{jiao2021new}, we construct both an image dataset (sourced from LVIS~\cite{gupta2019lvis}) and a video dataset (sourced from Ego4D~\cite{grauman2022ego4d}) as part of \dataname{}. Overall, \dataname{} has $641K$ part masks annotated in $77K$ images for $260K$ object instances across $75$ object classes and $456$ object-specific part classes. It has an order of magnitude more objects with parts, compared to recently introduced PartImageNet dataset~\cite{he2021partimagenet}. \dataname{} further provides annotations for $55$ different attributes for both objects and parts. We conducted user studies and multi-round manual curation to identify high-quality vocabulary of parts and attributes.


Along with the dataset, we provide three associated benchmark tasks to help the community evaluate its progress over time. These tasks include: a) part segmentation, b) attribute detection for objects and object-parts and c) zero-shot instance detection with part/attribute queries. The first two tasks are aimed at benchmarking stand alone capabilities of part and attribute understanding. The third task evaluates models directly for a downstream task.


While building the dataset and benchmarks, we navigate some key design choices: (a) Should we evaluate parts and attributes conditioned on the object or independent of the objects (eg: evaluating “leg” vs. “dog-leg”, “red” vs. “red cup”)? (b) How do we keep annotation workload limited without compromising fair benchmarking? 

To answer the first question, we observed that the same semantic part can visually manifest very differently in different objects (``dog-leg” vs ``chair-leg”). This makes the parts of different objects virtually independent classes, prompting us to evaluate them separately. This also forces models to not just identify parts or attributes independently, but predict objects, parts and attributes jointly. This is more useful for downstream applications.

Next, to keep annotation costs limited, we can construct a federated dataset as suggested in LVIS~\cite{gupta2019lvis}. For object detection, LVIS showed that this enables fair evaluation without needing exhaustive annotations for every image. However, this poses a specific challenge in our setup. Object detection requires every region to be associated with only one label (object category), while we require multiple labels: object, part and attribute jointly. This subtle but important difference, makes it non-trivial to extend definition and implementation of metrics from LVIS to our setup. We provide a nuanced treatment of missing labels at different levels (missing attribute labels vs. missing part and attribute labels) to handle this. 

Our design choices allow us to use popular detection metrics: Average Precision and Average Recall for all our tasks. To facilitate calibration of future research models, we also provide benchmark numbers for all tasks using simple variants of mask R-CNN~\cite{he2017mask} and ViT-det~\cite{Li2022ExploringPV}. Dataset, models, and code are open-sourced at \href{https://github.com/facebookresearch/paco}{https://github.com/facebookresearch/paco}.



%% file: related.tex
\begin{table*}
    \setlength{\tabcolsep}{1.5mm}  
    \centering
    \footnotesize
    \begin{tabular}{c||c|c|c|c|c|c|c||c|c|c}
    \hline
    & PartsIN & Pascal & City.-PP & VAW & COCO att. & FashionPedia & ADE & \datalvis{} & \dataego{} & \dataname{} \\
    \hline \hline
   object domain & comm. & comm. & comm. & comm. & comm. & fashion & comm. & comm. & comm. & \textbf{comm.} \\
    \# obj cats & 158 & 20 & 5 & 2260 & 29 & 27 & 2693 & 75 & 75 & 75 \\
    \# img with obj mask & 24K & 20K & 3.5K & 72.3K & 84K & 48.8K & 27.6K & 57.6K & 23.9K & 81.5K \\
    \# obj mask & 24K & 50k & 56k & 260.9K & 180K & 167.7K	& 434.8K & 274K & 58.4K & 332.3K \\
    \hline
    \# obj-part cats & 609 & 193 & 23 & - & - & - & 476 & 456 & 456 & 456 \\
    \# obj-agn. part cats & 13 & 127 & 9 & - & - & 19 & - & 200 & 194 & \textbf{200} \\
    \# img with part mask & 24K & 19K &	3.5K & - & - & 48.8K & 12.6K & 52.7K & 24K & \textbf{76.7K} \\
    \# part mask & 112K & 363.5k & 100k & -  & - & 174.4K & 193.2K & 502K & 139.3K & \textbf{641.4K} \\
    \# obj with part mask & 24K	& 40k & 31k & - & -	& NA & 59K & 209.4K & 50.9K & \textbf{260.3K} \\
    \hline
    \# att cats & -	& - & - & 620	& 196 & 294 & 1314 & 55 & 55 & 55 \\
    \# img with att & - & - & - & 72.3K	& 84K	& 48.8K	& 16.3K & 48.6K &	26.3K & \textbf{74.9K} \\
    \# obj with att & - & - & - & 260.9K & 180K & 78.9K & 74.6K	& 74.4K & 49.6K & 124K \\
    \# part with att & - & - & - &	- & - & 132.8K & 31.4K & 186K & 110.6K & \textbf{296.6K} \\ 
    avg \# att / img & - & - & - & 3.6 & 41 & 8.4 & 24.7 & 22.2 & 25.8 & 23.4 \\
    neg. att labels & - & - & - & TRUE & TRUE & TRUE & FALSE & TRUE & TRUE & \textbf{TRUE} \\ \hline
    
    \end{tabular}
    \caption{Comparison of publicly available parts and attributes datasets. PartsIN refers to PartsImageNet, City.-PP refers to Cityscape PanopticParts. Salient features of our dataset are shown in bold.}
    \label{tab:dataset_stats}
\end{table*}

\subsection{Related work}
Availability of large-scale datasets like ImageNet~\cite{deng2009imagenet}, COCO~\cite{lin2014microsoft}, LVIS~\cite{gupta2019lvis} have played a crucial role in the acceleration of object understanding. We briefly review datasets that provide a variety of annotations for objects besides category labels.




{\noindent\textbf{Object detection and segmentation datasets }}\\
The task of detecting and segmenting object instances is well studied with popular benchmark datasets such as COCO~\cite{lin2014microsoft}, LVIS~\cite{gupta2019lvis}, Object365~\cite{shao2019objects365}, Open Images~\cite{kuznetsova2020open} and Pascal~\cite{Everingham15} for common objects. There are also domain-specific datasets for fashion~\cite{jia2020fashionpedia,zheng2018modanet}, medical images~\cite{yan2019mulan} and OCR~\cite{deng2012mnist,veit2016coco,singh2021textocr}. Recent datasets like LVIS, OpenImages and Objects365 have focused on building larger object-level vocabulary without specific focus on parts or attributes. In particular, LVIS introduced the idea of federated annotations, making it possible to scale to larger vocabularies without drastically increasing annotation costs. We adopt this in our dataset construction as well.

{\noindent\textbf{Part datasets }}\\
Pixel-level part annotations for common objects are provided by multiple datasets such as PartImageNet~\cite{he2021partimagenet}, PASCAL-Part~\cite{chen2014detect}, ADE20K~\cite{zhou2017scene,zhou2019semantic} and Cityscapes-Panoptic-Parts~\cite{meletis2020cityscapes}. PASCAL provides part annotations for $20$ object classes and PartImageNet provides parts for animals, vehicles and bottle. Cityscapes has parts defined for $9$ object classes. In contrast we focus on a larger set of $75$ common objects from LVIS vocabulary. Our dataset has ten times larger number of object boxes annotated with part masks compared to PartImageNet. ADE20K is a $28K$ image dataset for scene parsing which includes part masks. While it provides an instance segmentation benchmark for $100$ object categories, part segmentation is benchmarked only for $8$ object categories due to limited annotations. We provide a part segmentation benchmark for all $75$ object classes. More detailed comparison of above datasets are provided in Tab.~\ref{tab:dataset_stats}. Apart from common objects, part segmentation has also been studied for specific domains like human part segmentation: LIP~\cite{gong2017look}, CIHP~\cite{yang2019parsing}, MHP~\cite{li2017multiple}, birds: CUB-200~\cite{wah2011caltech}, fashion: ModaNet~\cite{zheng2018modanet}, Fashionopedia and cars: CarFusion~\cite{Reddy_2018_CVPR}, ApolloCar3D~\cite{song2019apollocar3d}.

{\noindent\textbf{Attribute datasets }}\\
Attributes have long been viewed as a fundamental way to describe objects. In particular, domain-specific attribute datasets have become more prevalent for fashion, animals, people, faces and scenes~\cite{guo2019imaterialist,kosti2017emotion,li2016human,liu2016deepfashion,zhou2017places,zhou2019semantic}. A motivation of our work is to extend such rich descriptions to common objects and object parts as well. More recently, Pham et al.~\cite{pham2021learning} introduced  the Visual Attributes in the Wild (VAW) dataset constructed from two source datasets: VGPhraseCut~\cite{wu2020phrasecut} and GQA~\cite{hudson2019gqa}. VAW expanded and cleaned the attributes in the source datasets, and adds explicit negative attribute annotations to provide a rigorous benchmark for object attribute classification. VAW solely focused on attribute classification, and assumed the object box and label to be known apriori. VAW is not benchmarked for joint end-to-end object/part localization and attribute recognition, which is the focus of our work.

{\noindent\textbf{Part and attribute datasets }}\\
Fashionpedia~\cite{jia2020fashionpedia} is a popular dataset for fashion providing both part and attribute annotations in an image. It is the closest line of work that also provides part localization and attribute recognition benchmarks. \dataname{} aims to generalize this to common object categories.

{\noindent\textbf{Instance recognition with queries }}\\
Attributes have been long used for zero-shot object recognition~\cite{romera2015embarrassingly,xian2017zero}. We use this observation to build an instance-level retrieval benchmark for retrieving a specific instance of an object from a collection of images using part and attribute queries. Recently, Cops-Ref~\cite{Chen_2020_CVPR} also introduced a challenging benchmark for object retrieval in the natural language setting with a focus on referring expressions~\cite{kazemzadeh2014referitgame,mao2016generation} that involve spatial relationships between objects. \dataname{} is aimed at benchmarking part and attribute based queries at varying levels of compositions.

%% file: dataset_construction.tex

\section{Dataset construction}
\subsection{Image sources}
\dataname{} is constructed from LVIS~\cite{gupta2019lvis} in the image domain and Ego4D~\cite{grauman2022ego4d} in the video domain. We chose LVIS due to its large object vocabulary and federated dataset construction. Ego4D has temporally aligned narrations, making it easy to source frames corresponding to specific objects.

\subsection{Object vocabulary selection}
 We first mined all object categories mentioned in the narrations accompanying Ego4D and took the intersection with common and frequent categories in LVIS. We then chose categories with at-least $20$ instances in Ego4D, resulting in $75$ categories commonly found in both LVIS and Ego4D.

\subsection{Parts vocabulary selection}
Excluding specific domains like fashion~\cite{jia2020fashionpedia}, there is no exhaustive ontology of parts for common objects. We mined part names from web-images obtained through queries like ``parts of a car". These images list part-names along with illustrations and pointers to the parts in the object. We manually curate such mined part names for an object category to only retain parts that are visible in majority of the object instances and clearly distinguishable. More details in the \suppref{}. This resulted in a total of $200$ part classes shared across all $75$ objects. When expanded to object-specific parts this results in $456$ object-part classes.




\subsection{Attribute vocabulary selection}
Attributes are particularly useful in distinguishing different instances of the same object type. Motivated by this, we conducted an in-depth user study (details in \suppref{}) to identify the sufficient set of attributes that can separate all object instances in our dataset. This led to the final vocabulary of $29$ colors, $10$ patterns and markings, $13$ materials and $3$ levels of reflectance.

\subsection{Annotation pipeline}
Our overall data annotation pipeline consists of: a) Object bounding box and mask annotation (only for Ego4D) b) part mask annotation, c) object and part attributes annotation and d) instance IDs annotation (only for Ego4D).

\subsubsection{Object annotation}
Bounding boxes and masks are already available for the $75$ object classes in LVIS, but not in Ego4D. For Ego4D, we use the provided narrations to identify timestamps in videos for specific object classes. We sampled $100$ frames around these timestamps and asked annotators to choose at most $5$ diverse (in lighting, viewpoint, etc.) frames that depict an instance of the object class. These frames are annotated with bounding boxes and object masks. A frame annotated with a specific object class is exhaustively annotated with every bounding box of the object class. For each object class in the evaluation splits we annotate negative images that are guaranteed to not contain the object.


\subsubsection{Part mask annotation}
We provide part masks for all annotated object boxes in both LVIS and Ego4D.
A fraction of the object boxes were rejected by annotators due to low resolution, motion blur or significant occlusion. This resulted in a total of $209K$, $43K$ object boxes with parts in LVIS, Ego4D respectively. For an object box to be annotated, we listed all the potential parts for the object class and asked annotators to annotate masks for the visible parts. Note that parts can be overlapping (for example, door and handle). We do not distinguish between different instances of a part in an object instance, but provide a single mask covering all pixels of a part class in the object (e.g., all car wheels are covered by a single mask).


\subsubsection{Attributes annotation}
Every bounding box in Ego4D is annotated with object and part-level attributes, unless rejected by annotators due to lack of resolution or blur. Obtaining exhaustive attribute annotations for all object and part instances in LVIS dataset for the $75$ categories is very expensive. Hence, we randomly selected one medium or large\footnote{Decided based on box area as defined in COCO~\cite{veit2016coco}.} bounding box per image, per object class for attribute annotations.
We annotate a box with both object-level and part-level attributes for all $55$ attributes in a single annotation job. This ensures consistency between object and part attributes and helped us annotate attributes for a diverse set of images with limited expense.
This resulted in $74K$ ($50K$) object instances and $186K$ ($111K$) part instances annotated with attributes for LVIS (Ego4D) respectively. 





\subsubsection{Instance annotation}
We also introduce a zero-shot instance detection task with our dataset. To do this we need unique instance IDs for each object box in the dataset. For LVIS data, we assume each individual object box to be a separate instance. However, this is not true for Ego4D. Different bounding boxes of an object could correspond to the same instance. Also, different videos in Ego4D could have the same object instance. We underwent a rigorous multi-stage process to annotate instance IDs, explained in the \suppref{}. This resulted in $16908$ unique object instances among the $49955$ annotated object boxes in Ego4D.


\subsubsection{Managing annotation quality}
Each stage in the annotation pipeline had multiple associated quality control methods such as use of gold standard and annotation audits. We had $10-50$ instances of each object annotated by expert annotators and set aside as gold annotations. For part mask annotations, we measured mIoU with gold images for each object class and re-annotated object classes with mIoU $<50\%$ on gold annotations. Eventually, $90\%$ of the object classes have mIoU $\ge 0.75$ with the gold-annotated masks (shown in \suppref{}). For all attribute annotations we were checking quality by randomly sampling annotations, finding patterns in annotation errors, updating guidelines to correct clear biases, and re-annotating erroneous examples. This eventually drove accuracy to more than $85\%$ on the gold annotations provided by expert annotators.



%% file: dataset_stats.tex
\section{Dataset statistics}\label{sec:dataset_stats}

\begin{figure}
     \centering
     \begin{subfigure}[b]{0.48\columnwidth}
         \centering
         \includegraphics[width=\linewidth]{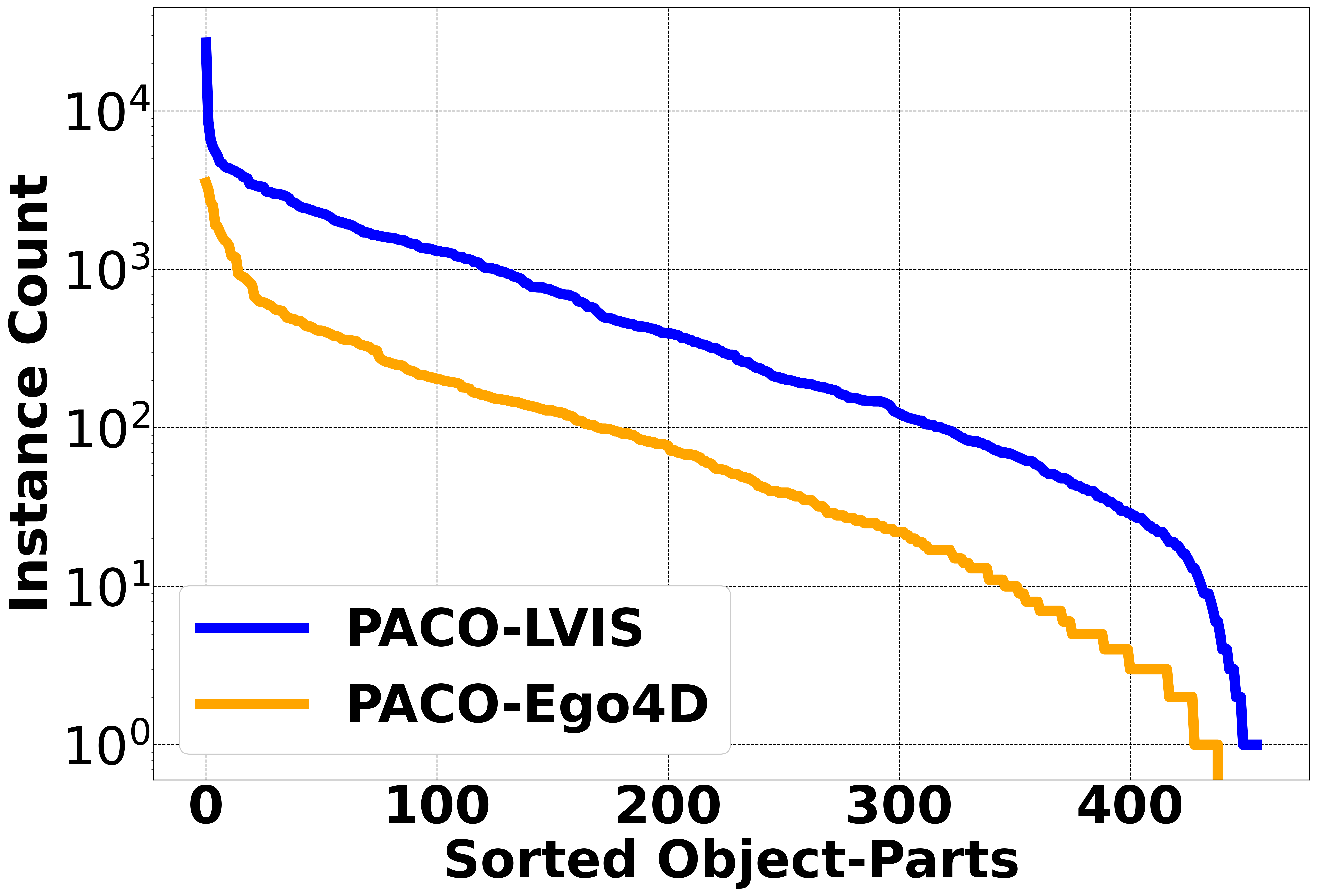}
         \caption{}
         \label{fig:Number of instances per object-part category}
     \end{subfigure}
     \begin{subfigure}[b]{0.48\columnwidth}
         \centering
         \includegraphics[width=\linewidth]{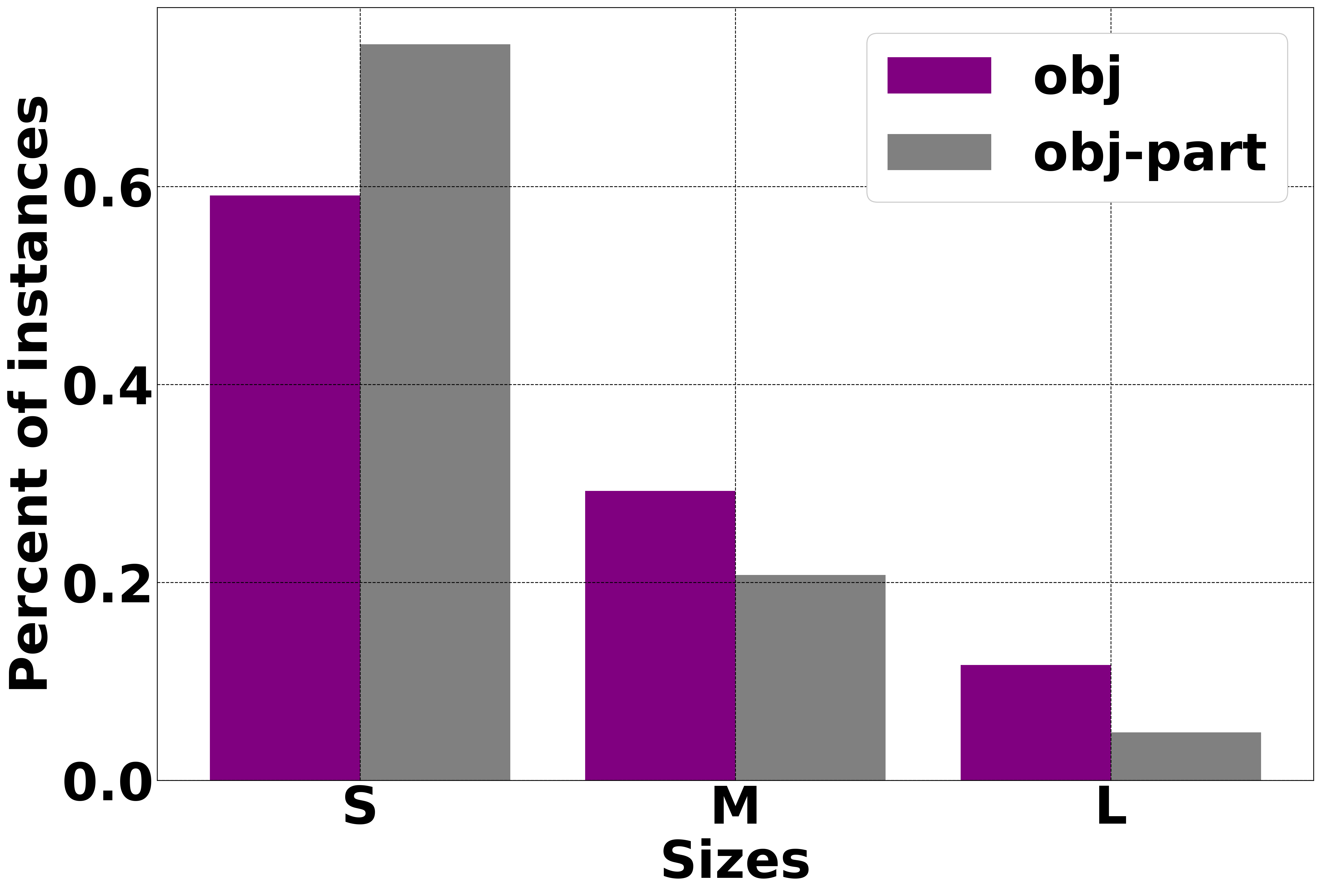}
         \caption{}
         \label{fig:Distribution of sizes across instances}
     \end{subfigure}
     \begin{subfigure}[b]{\columnwidth}
         \centering
         \includegraphics[width=\linewidth]{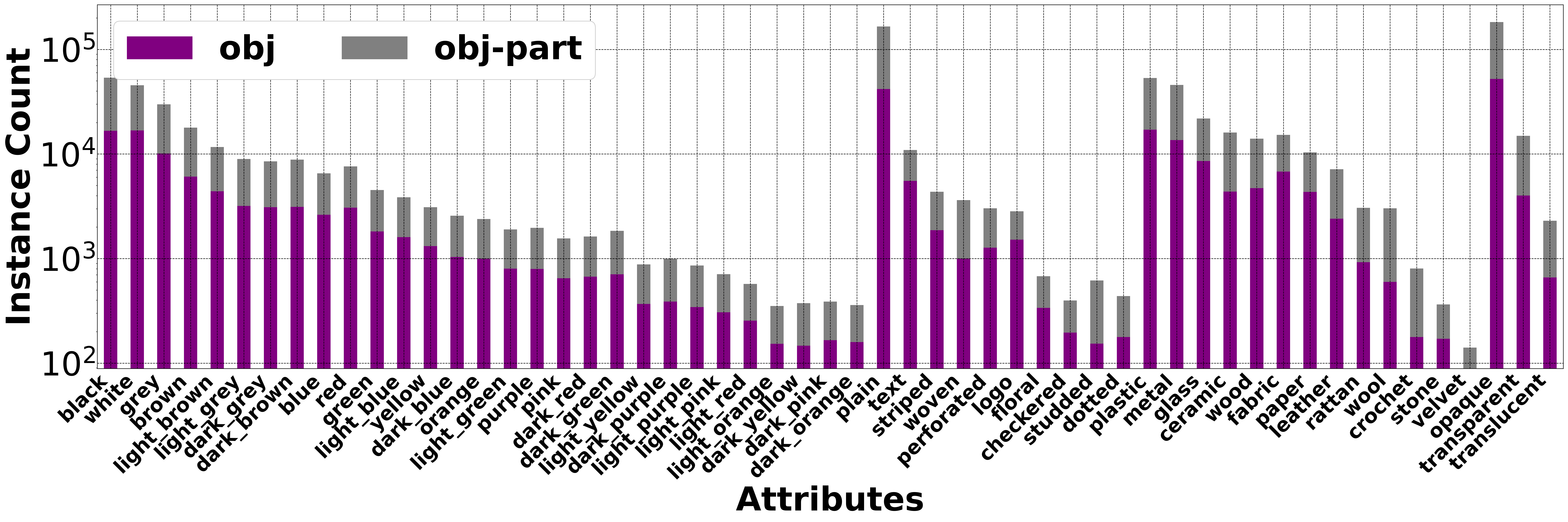}
         \caption{}
         \label{fig:Distribution of attribute classes across instances}
     \end{subfigure}
     \caption{Dataset Statistics. Fig. (a) shows the distribution of instances across the $456$ object-part categories. Fig. (b) shows the size distribution of object and part masks in \datalvis{}. Fig. (c) shows the distribution of the {55} attribute classes across instances in \datalvis{}}
    \label{fig:Dataset Statistics}
\end{figure}




{\noindent{\textbf{Part statistics:}}}
Fig.~\ref{fig:Number of instances per object-part category} shows the number of part masks annotated for each object-part category in \datalvis{} and \dataego{}. We observe the typical long-tail distribution with certain categories like `book-cover', `chair-back' and `box-side' having greater than $6500$ instances, and, categories like `fan-logo' and `kettle-cable' having fewer than $5$ instances. Fig.~\ref{fig:Distribution of sizes across instances} shows the distribution of number of large, medium and small parts in \datalvis{}. We observe that larger fraction of part masks belong to low and medium size, compared to object masks. 



{\noindent{\textbf{Attribute statistics: }}}
Fig.~\ref{fig:Distribution of attribute classes across instances} shows number of annotations per attribute and attribute type in \datalvis{}. We again observe a long-tail distribution with common attributes like colors having many annotations, while uncommon ones like `translucent' having fewer annotations. 




{\noindent{\textbf{Comparison with other datasets: }}}
We also provide an overview of different parts and/or attributes datasets in Tab.~\ref{tab:dataset_stats}. Among the datasets with part annotations, \dataname{} provides $641K$ part mask annotations in the joint dataset, which is $~3\times$ bigger than other datasets like ADE20K ($176K$), PartImageNet ($112K$) and Fashionpedia ($175K$). While ADE20K has sizeable number of part masks overall, it doesn't provide a well defined instance-level benchmark for parts due to limited test annotations. \dataname{} has $10\times$ more object instances with parts ($260K$) compared to the next closest parts benchmark dataset for common objects: PartsImageNet ($25K$). In terms of attributes, the joint dataset has $124K$ object and $297K$ part masks with attribute annotations. While VAW has $261K$ object masks with attributes, the combined set of attribute annotations for part and object masks ($421K$) in \dataname{} is still larger. VAW  has a larger vocabulary of attributes $620$ vs $55$. However, in \dataname{}, every object/part mask annotated with attributes is exhaustively annotated with all attributes in the vocabulary unlike VAW. This makes the density of attributes per image $23.4$ much larger than VAW $3.6$. COCO-attributes provides attribute annotations for COCO images as well, but for much smaller set of object classes ($29$).


%% file: evaluation_benchmark.tex
\section{Tasks and evaluation benchmark}
We now introduce three evaluation tasks. Our first two tasks directly evaluate the quality of parts segmentation and attributes prediction. The other task aims to leverage parts and attributes for zero-shot object instance detection.

\subsection{Dataset splits} We split both \datalvis{} and \dataego{} datasets into \texttt{train}, \texttt{val} and \texttt{test} sets. The \texttt{test} split of \datalvis{} is a strict subset of the LVIS-v1 \texttt{val} split and contains $9443$ images. The \texttt{train} and \texttt{val} splits of \datalvis{} are obtained by randomly splitting LVIS-v1 \texttt{train} subset for $75$ classes, and contain $45790$ and $2410$ images respectively. Ego4D is split into $15667$ \texttt{train}, $825$ \texttt{val} and $9892$ \texttt{test} images. The set of object instance IDs in Ego4D \texttt{train} and \texttt{test} sets are disjoint. 

\subsection{Federated dataset for object categories} We briefly review the concept of federated dataset from LVIS~\cite{gupta2019lvis}, where every image in the evaluation set is not annotated exhaustively with all object categories. However, every object category has (a) a set of \emph{negative images} that are guaranteed to not contain any instance of the object, (b) a set of \emph{exhaustive positive images} where all instances of the object are annotated and (c) a set of \emph{non-exhaustive positive images} with at-least one instance of the object annotated. Non-exhaustive positive images are not guaranteed to have all instances of the object annotated. Only these three types of images are used to evaluate AP for the category.

\subsection{Part segmentation}
Our part segmentation task requires an algorithm to detect and segment the part-masks of different object instances in an unseen image and assign an $(object,part)$ label with a confidence score to the part-mask. The $(object,part)$ pairs are from a fixed know set. This is similar to the object instance segmentation task, but uses object-part labels instead of only object labels. We consider parts of different instances of the object in an image to be different object-part instances.

We choose to evaluate the task for $(object,part)$ labels instead of only $part$ labels, since the appearance and definition of the same semantic part can be very different depending on the object it appears in. We expect the models to produce both an object and a part label, with a single joint score. This leaves us with $456$\footnote{Similar to LVIS, a small number of valid $(object,part)$ pairs in \texttt{train} do not have any annotated instances in the \texttt{val} and \texttt{test} splits. We ignore these object-parts for evaluation.} object-parts in the dataset.

We use mask and box Average Precision (AP) metrics defined in COCO~\cite{veit2016coco}. AP is averaged over different thresholds of intersection over union (IoU)\footnote{Mask IoU is used for mask AP and box IoU is used for box AP~\cite{veit2016coco}}. 


{\noindent\textbf{AP calculation in federated setup }}\\
Given a set of predicted masks with a combined score for (object category $o$, part category $p$), we compute AP for the object-part $(o,p)$ at a given IoU threshold. We use all positive and negative images of $o$ to do this. We treat each predicted mask as a \emph{true positive}, \emph{false positive} or \emph{ignore} it based on the following criteria.

{\noindent\textbf{Negative images: }} We treat all predicted masks in negative images of object $o$ as \emph{false positives} for the object-part $(o,p)$. This is a valid choice, since an object-part cannot be present without the object.

{\noindent\textbf{Non-exhaustive positive images: }} We treat images marked as non-exhaustive for the object category as non-exhaustive for the object-part as well. There is also a subset of images exhaustively annotated for the object, but not for the object-part. We provide an explicit flag to identify such additional non-exhaustive images for every object-part in our datasets. In both cases of non-exhaustive images, we consider predicted masks overlapping (above the IoU threshold) with an annotated ground-truth object-part mask as \emph{true positives}. We \emph{ignore} other predicted masks in the images.

{\noindent\textbf{Exhaustive positive images: }} On the subset of positive images, where every instance of the object-part is exhaustively annotated, we treat predicted masks as \emph{true positives} if they overlap (above the threshold) with a ground-truth annotated part mask, otherwise they are treated as \emph{false positives}. 

The true and false positive masks along with their predicted scores are used to calculate AP at a given threshold as defined in COCO~\cite{veit2016coco}. We report mean Average Precision across all object-part categories ($AP^{opart}$).


\begin{table}
    \setlength{\tabcolsep}{2mm}  
    \centering
    \footnotesize
    \begin{tabular}{c||c c|c c}
    \hline
    & \multicolumn{2}{|c}{mask $AP$} & \multicolumn{2}{|c}{box $AP$} \\
    Model & \apobj & \apopart & \apobj & \apopart\\
    \hline \hline
    R50 FPN & 31.5 \deemph{$\pm$ 0.3} & 12.3 \deemph{$\pm$ 0.1} & 34.6 \deemph{$\pm$ 0.3} & 16.0 \deemph{$\pm$ 0.1} \\
   + cascade & 32.6 \deemph{$\pm$ 1.3} & 12.5 \deemph{$\pm$ 0.7} & 37.4 \deemph{$\pm$ 1.6} & 16.3 \deemph{$\pm$ 1.1} \\
    \hline
    R101 FPN & 31.5 \deemph{$\pm$ 0.6} & 12.3 \deemph{$\pm$ 0.3} & 34.8 \deemph{$\pm$ 0.8} & 16.1 \deemph{$\pm$ 0.3} \\
   + cascade & 35.1 \deemph{$\pm$ 0.1} & 13.7 \deemph{$\pm$ 0.1} & 40.2 \deemph{$\pm$ 0.1} & 17.9 \deemph{$\pm$ 0.2} \\
    \hline
    ViT-B FPN & 33.6 \deemph{$\pm$ 0.3} & 13.5 \deemph{$\pm$ 0.1} & 38.7 \deemph{$\pm$ 0.4} & 17.5 \deemph{$\pm$ 0.0} \\
    + cascade & 33.6 \deemph{$\pm$ 0.3} & 13.5 \deemph{$\pm$ 0.1} & 38.7 \deemph{$\pm$ 0.4} & 17.5 \deemph{$\pm$ 0.0} \\
    \hline
    ViT-L FPN & 42.8 \deemph{$\pm$ 0.3} & 17.3 \deemph{$\pm$ 0.1} & 47.3 \deemph{$\pm$ 0.2} & 22.0 \deemph{$\pm$ 0.1} \\
    + cascade & 43.4 \deemph{$\pm$ 0.3} & 17.7 \deemph{$\pm$ 0.0} & 49.7 \deemph{$\pm$ 0.2} & 22.9 \deemph{$\pm$ 0.0} \\
    \hline
    \end{tabular}
    \caption{Object and object-part segmentation results for mask-RCNN and ViT-det models trained and evaluated on PACO-LVIS}
    \label{tab:part_seg_results}
\end{table}

\subsection{Instance-level attributes prediction}
In \dataname{}, this is the task that requires an algorithm to produce masks and/or boxes along with both a category label (object or object-part) as well as an attribute label and a single joint confidence score for the category with the attribute (eg.: score for ``red car", ``red car-wheel").



Since multiple aspects are being evaluated together, we need to be meticulous in designing the evaluation metric. In particular, we need to be careful in our consideration of object and object-part masks with missing attribute annotations as we show next.

\begin{table*}
    \setlength{\tabcolsep}{1.5mm}  
    \centering
    \footnotesize
    \begin{tabular}{c || c | c c c c || c | c c c c }
    \hline
    Model & $AP^{obj}_{att}$ & $AP^{obj}_{col}$ & $AP^{obj}_{pat}$ & $AP^{obj}_{mat}$ & $AP^{obj}_{ref}$ & $AP^{opart}_{att}$ & $AP^{opart}_{col}$ & $AP^{opart}_{pat}$ & $AP^{opart}_{mat}$ & $AP^{opart}_{ref}$ \\
    \hline \hline
    R50 FPN & 13.5 \deemph{$\pm$ 0.3} & 10.8 \deemph{$\pm$ 0.1} & 14.1 \deemph{$\pm$ 0.6} & 9.9 \deemph{$\pm$ 0.4} & 19.1 \deemph{$\pm$ 0.7} & 9.7 \deemph{$\pm$ 0.2} & 10.7 \deemph{$\pm$ 0.2} & 10.6 \deemph{$\pm$ 0.5} & 6.9 \deemph{$\pm$ 0.0} & 10.7 \deemph{$\pm$ 0.2} \\
     + cascade & 15.0 \deemph{$\pm$ 1.0} & 12.4 \deemph{$\pm$ 0.7} & 16.1 \deemph{$\pm$ 0.7} & 11.0 \deemph{$\pm$ 0.9} & 20.6 \deemph{$\pm$ 1.6} & 10.5 \deemph{$\pm$ 0.7} & 11.6 \deemph{$\pm$ 0.8} & 11.6 \deemph{$\pm$ 0.8} & 7.6 \deemph{$\pm$ 0.7} & 11.2 \deemph{$\pm$ 0.7} \\
    \hline
      R101 FPN & 13.5 \deemph{$\pm$ 0.3} & 11.0 \deemph{$\pm$ 0.2} & 13.9 \deemph{$\pm$ 0.3} & 9.9 \deemph{$\pm$ 0.4} & 19.1 \deemph{$\pm$ 0.6} & 9.9 \deemph{$\pm$ 0.1} & 11.0 \deemph{$\pm$ 0.4} & 10.8 \deemph{$\pm$ 0.4} & 7.1 \deemph{$\pm$ 0.2} & 10.9 \deemph{$\pm$ 0.3} \\
     + cascade & 16.0 \deemph{$\pm$ 0.1} & 13.4 \deemph{$\pm$ 0.2} & 16.7 \deemph{$\pm$ 0.2} & 12.3 \deemph{$\pm$ 0.1} & 21.5 \deemph{$\pm$ 0.4} & 11.5 \deemph{$\pm$ 0.2} & 12.6 \deemph{$\pm$ 0.1} & 12.5 \deemph{$\pm$ 0.3} & 8.5 \deemph{$\pm$ 0.3} & 12.6 \deemph{$\pm$ 0.3} \\
    \hline
      ViT-B FPN & 15.0 \deemph{$\pm$ 0.2} & 11.9 \deemph{$\pm$ 0.1} & 14.9 \deemph{$\pm$ 0.5} & 12.8 \deemph{$\pm$ 0.4} & 20.4 \deemph{$\pm$ 0.8} & 10.9 \deemph{$\pm$ 0.2} & 11.3 \deemph{$\pm$ 0.3} & 11.4 \deemph{$\pm$ 0.6} & 9.0 \deemph{$\pm$ 0.1} & 11.8 \deemph{$\pm$ 0.3} \\
     + cascade & 15.7 \deemph{$\pm$ 0.2} & 12.6 \deemph{$\pm$ 0.1} & 16.0 \deemph{$\pm$ 0.5} & 13.2 \deemph{$\pm$ 0.4} & 20.9 \deemph{$\pm$ 0.5} & 11.0 \deemph{$\pm$ 0.2} & 11.6 \deemph{$\pm$ 0.2} & 11.7 \deemph{$\pm$ 0.4} & 9.0 \deemph{$\pm$ 0.2} & 11.5 \deemph{$\pm$ 0.3} \\
    \hline
      ViT-L FPN & 18.8 \deemph{$\pm$ 0.3} & 14.9 \deemph{$\pm$ 0.2} & 18.9 \deemph{$\pm$ 1.0} & 16.0 \deemph{$\pm$ 0.7} & 25.4 \deemph{$\pm$ 0.7} & 13.5 \deemph{$\pm$ 0.2} & 14.0 \deemph{$\pm$ 0.2} & 14.0 \deemph{$\pm$ 0.4} & 11.7 \deemph{$\pm$ 0.4} & 14.3 \deemph{$\pm$ 0.6} \\
     + cascade & 19.5 \deemph{$\pm$ 0.3} & 15.6 \deemph{$\pm$ 0.3} & 19.1 \deemph{$\pm$ 0.5} & 16.3 \deemph{$\pm$ 0.3} & 27.0 \deemph{$\pm$ 0.4} & 13.8 \deemph{$\pm$ 0.1} & 14.4 \deemph{$\pm$ 0.3} & 15.1 \deemph{$\pm$ 0.0} & 11.5 \deemph{$\pm$ 0.2} & 14.5 \deemph{$\pm$ 0.4} \\
    \end{tabular}
    \caption{Attribute prediction results for mask R-CNN and ViT-det models trained and evaluated on \datalvis{}. Box $AP$ results are shown for both object attributes and object-part attributes prediction.}
    \label{tab:attr_ressults}
\end{table*}

\begin{table}
    \setlength{\tabcolsep}{2mm}  
    \centering
    \footnotesize
    \begin{tabular}{c || c c c }
    Model & LB-no attribute & Original & UB-perfect attribute \\
    \hline \hline
    R-50 FPN & 8.6 \deemph{$\pm$ 0.3} & 13.5 \deemph{$\pm$ 0.3} & 61.4 \deemph{$\pm$ 0.3} \\
    R-101 FPN & 8.6 \deemph{$\pm$ 0.3} & 13.5 \deemph{$\pm$ 0.3} & 63.0 \deemph{$\pm$ 0.3} \\
    ViT-B FPN & 9.0 \deemph{$\pm$ 0.1} & 15.0 \deemph{$\pm$ 0.2} & 60.5 \deemph{$\pm$ 0.1} \\
    ViT-L FPN & 10.6 \deemph{$\pm$ 0.2} & 18.8 \deemph{$\pm$ 0.3} & 72.6 \deemph{$\pm$ 0.3} \\
    \end{tabular}
    \caption{Bounds for $AP^{obj}_{att}$ keeping detection quality fixed and changing attribute scores. For lower bound (LB), we neglect attribute scores and for upper bound (UB), we assume perfect attribute scores.}
    \vspace{-4mm}
    \label{tab:attr-AP-bounds}
\end{table}

{\noindent\textbf{AP calculation in federated setup }}\\
We continue with AP as our evaluation metric. Given a set of predicted masks with scores for a category $c$ (can be an object $o$ or object-part $(o,p)$) and attribute $a$ combination, we compute AP for $(c,a)$. We use all positive or negative images of object $o$ to compute the AP for $(c,a)$. We compute AP at different IoU thresholds and report the average. At a given threshold, we identify true positives, false positives or ignored masks as described below.

{\noindent\textbf{Negative images:}} We treat all predicted masks in negative images of the object $o$ as \emph{false positives} for $(c,a)$.

{\noindent\textbf{Positive images: }} In both exhaustive and non-exhaustive positive images, we do the following. We treat masks overlapping with ground-truth masks of the category that are also annotated positively for the attribute $a$ as \emph{true positives}. Masks overlapping with ground truth masks of the category $c$, but annotated negatively for attribute $a$ are treated as \emph{false positives}. We \emph{ignore} mask predictions that overlap with ground-truth masks of category $c$ with un-annotated attribute labels. We differ in the treatment of mask predictions not overlapping with any ground-truth mask of the category, in exhaustive and non-exhaustive positive images. In case of non-exhaustive images, we \emph{ignore} such predictions, while in exhaustive images we treat such predictions as \emph{false positives}. 

We use the true and false positives along with their predicted confidence scores to calculate AP for $(c,a)$. We only compute AP for $(c,a)$ if at-least one instance of $c$ is positively annotated with attribute $a$ in test set and at-least $40$ other instances of $c$ are negatively annotated for $a$.


We observe that some $(c,a)$ combinations can be ``rare" in the evaluation set with few positive occurrences only. As observed in LVIS~\cite{gupta2019lvis} such ``rare" combinations can have a higher variance in the metric and it helps to average the metric across categories to reduce variance. Hence, we aggregate AP at an attribute level for $a$, by averaging the AP across all categories that are evaluated with $a$. We aggregate over object categories and object-part categories separately, leading to object AP ($AP^{obj}_a$) and object-part AP ($AP^{opart}_a$) for each attribute $a$. In our experiments, we report the mean value of $AP^{obj}_a$ across all attributes: $AP^{obj}_{att}$, as well as the mean values across attributes belonging to color ($AP^{obj}_{col}$), pattern \& markings ($AP^{obj}_{pat}$), material ($AP^{obj}_{mat}$) and reflectance ($AP^{obj}_{ref}$). We do the same for object-parts and report $AP^{opart}_{att}$, $AP^{opart}_{col}$, $AP^{opart}_{pat}$, $AP^{opart}_{mat}$ and $AP^{opart}_{ref}$.

\subsection{Zero-shot instance detection}
Zero-shot instance detection requires an algorithm to retrieve the bounding box of a specific instance of an object based on a ``query" describing the instance. No sample images of the instance are previously seen by the algorithm. This has similarity to referring expression tasks~\cite{kazemzadeh2014referitgame,mao2016generation,Chen_2020_CVPR} that localize a specific object instance in an image based on attribute and spatial relation queries. However, we introduce a more fine-grained evaluation benchmark, where the queries are composed of both object and part attributes at different levels of composition.

We construct the evaluation dataset for both LVIS and Ego4D from their corresponding \texttt{test} splits. We first define level-k (L$k$) query as describing an object instance in terms of $k$ attributes of the object and/or parts. For example, "blue mug" or ``mug with a blue handle" are sample L1 queries, ``blue striped mug" is a L2 query and ``blue striped mug with white handle" is a L3 query. Each query is associated with $1$ positive image with a bounding box and a distractor set of up to $100$ images, see Fig~\ref{fig:pull_figure}.



To ensure pracitcal utility, we avoid queries with uninformative attributes like ``car with a black wheel" since all cars have black wheel and eliminate part names that are infrequently used in large multimodal datasets (PMD~\cite{singh2022flava}). The distractor images for each query contain hard-negatives corresponding to other instances of the same object category, but differing by at-least one attribute from the query. Queries have more than $40\%$ hard negatives on average. \datalvis{} has $931/2348/2000$ and \dataego{} has $793/1437/2115$ L1/L2/L3 queries respectively.


We measure performance of an algorithm through average recall metrics $AR@k$ where $k=1,5$ denotes the top-k boxes returned by the method for a query. We compute AR at different IoU thresholds and report the average over all thresholds, as defined in COCO~\cite{veit2016coco}.

%% file: experiments.tex
\begin{table*}
    \centering
    \footnotesize
    \begin{tabular}{c || c c || c c || c c || c c }
  & \multicolumn{2}{c||}{L1 queries} & \multicolumn{2}{c||}{L2 queries} & \multicolumn{2}{c||}{L3 queries} & \multicolumn{2}{c}{all queries}\\
    \hline
    Model & $AR@1$ & $AR@5$ & $AR@1$ & $AR@5$ & $AR@1$ & $AR@5$ & $AR@1$ & $AR@5$ \\
    \hline \hline
    R50 FPN & 22.5 \deemph{$\pm$ 0.7} & 39.2 \deemph{$\pm$ 0.5} & 20.1 \deemph{$\pm$ 0.4} & 38.5 \deemph{$\pm$ 0.1} & 22.3 \deemph{$\pm$ 0.9} & 44.5 \deemph{$\pm$ 1.1} & 21.4 \deemph{$\pm$ 0.6} & 40.9 \deemph{$\pm$ 0.3} \\
    R101 FPN & 23.1 \deemph{$\pm$ 0.7} & 40.5 \deemph{$\pm$ 1.4} & 20.0 \deemph{$\pm$ 0.6} & 39.3 \deemph{$\pm$ 1.0} & 23.1 \deemph{$\pm$ 0.7} & 45.2 \deemph{$\pm$ 0.6} & 21.7 \deemph{$\pm$ 0.6} & 41.8 \deemph{$\pm$ 0.8} \\
      ViT-B FPN & 26.8 \deemph{$\pm$ 0.2} & 45.8 \deemph{$\pm$ 0.2} & 22.7 \deemph{$\pm$ 0.5} & 40.0 \deemph{$\pm$ 0.7} & 24.1 \deemph{$\pm$ 0.5} & 42.5 \deemph{$\pm$ 1.5} & 23.9 \deemph{$\pm$ 0.4} & 42.0 \deemph{$\pm$ 0.9} \\
      ViT-L FPN & 35.3 \deemph{$\pm$ 0.7} & 57.3 \deemph{$\pm$ 0.6} & 29.7 \deemph{$\pm$ 0.6} & 50.1 \deemph{$\pm$ 0.2} & 31.1 \deemph{$\pm$ 0.8} & 52.3 \deemph{$\pm$ 0.9} & 31.2 \deemph{$\pm$ 0.4} & 52.2 \deemph{$\pm$ 0.5} \\
    \end{tabular}
    \caption{Zero-shot instance detection results for different query levels for FPN models from Sec.~\ref{sec:attribute_exp} trained and evaluated on \datalvis{}.}
    \label{tab:zero_shot_inst_det}
\end{table*}

\begin{table}
    \centering
    \footnotesize
    \begin{tabular}{c || c | c | c }
    Model & $L1_{obj}$ & $L1_{part}$ & $L1$ \\
    \hline \hline
    MDETR R101 & 4.1 \deemph{$\pm$ 0.6} & 5.3 \deemph{$\pm$ 0.6} & 4.9 \deemph{$\pm$ 0.3} \\
    R101 FPN (Ours) & 20.3 \deemph{$\pm$ 0.9} & 24.4 \deemph{$\pm$ 1.0} & 23.1 \deemph{$\pm$ 0.7} \\
    \hline
    Detic Swin-B & 5.2 \deemph{$\pm$ 0.7} & 6.2 \deemph{$\pm$ 0.3} & 5.9 \deemph{$\pm$ 0.2} \\
    ViT-B FPN (Ours) & 22.6 \deemph{$\pm$ 0.8} & 28.9 \deemph{$\pm$ 0.6} & 26.8 \deemph{$\pm$ 0.2} \\
    \end{tabular}
    \caption{Zero-shot instance detection performance of open-vocabulary detectors on \datalvis{}. This is a difficult task for existing methods. We compare $AR@1$ on a subset of queries that are the closest to the detection task: $L1$ queries additionally split into subsets with only object ($L1_{obj}$) and only part ($L1_{part}$) attributes.}
    \label{tab:zero_shot_open_world}
\end{table}

\section{Benchmarking experiments}

\subsection{Part segmentation}
We train two mask R-CNN and two ViT-det~\cite{Li2022ExploringPV} models with $531$ classes comprising both $75$ object categories and $456$ object-part categories. We use the standard 100-epoch schedule recommended for LVIS with federated loss~\cite{zhu2020deformable} and LSJ~\cite{ghiasi2021simple} augmentation. For all experiments on part segmentation and attribute detection, we train on \texttt{train}, search for hyper-parameters on \texttt{val} and report results on \texttt{test} splits. More implementation details are in the \suppref{}. We trained with Cascade~\cite{cai2018cascade} as well as Feature Pyramid Network (FPN)~\cite{lin2017feature}. The results for models trained and evaluated on \datalvis{} are summarized in Tab.~\ref{tab:part_seg_results}. We also provide results for models trained on joint image + video \dataname{} dataset in the \suppref{}.


We observed that object-parts in general have a smaller AP compared to objects. This is due to the typically smaller size of parts compared to objects (Fig.~\ref{fig:Dataset Statistics}b). Nevertheless larger and better backbones like ViT-L are seen to improve performance for the part segmentation task. 

\subsection{Instance-level attributes prediction}
\label{sec:attribute_exp}
We train a simple extensions of mask R-CNN and ViT-det models with an additional attribute head on the shared backbone. The attribute head uses the same ROI-pooled features as the detection head to predict object and object-part attributes. We use a separate cross-entropy loss for each attribute type. The model is shown in more detail in the \suppref{}. We report box AP values for models trained on \datalvis{} in Tab.~\ref{tab:attr_ressults}. We also provide results for the joint dataset in the \suppref{}. During inference, we rank the detected boxes for a specific object-attribute combination by the product of the corresponding object and attribute scores. For parts, we rank boxes by product of corresponding object-part score and attribute score.


Attribute prediction is a much harder task than object detection, as witnessed by the lower AP values for both object-attributes and object-part-attributes, compared to object and part AP in Tab.~\ref{tab:part_seg_results}. We observe larger models fairing better for this task as well.

Since we measures multiple factors together, we analyze the sensitivity of $AP^{obj}_{attr}$ only to attribute prediction in Tab.~\ref{tab:attr-AP-bounds}. To do so, we keep detections from the trained models fixed and get (a) lower bounds by ignoring attribute scores and (b) upper bounds by assuming perfect attribute scores (details in \suppref{}). We observe a huge gap between lower and upper bounds, with our original models only partially bridging it. This shows scope for future improvements in the attribute prediction ability of the models.


\subsection{Zero-shot instance detection}
We generate benchmark numbers for this task by directly leveraging the models trained in Sec.~\ref{sec:attribute_exp}. For a given query, we use the scores corresponding to the object, part, object attributes, and part attributes mentioned in the query to rank object bounding boxes returned by the different joint models. We use a simple scoring function that combines these different scores using geometric mean to get one final score for each box (explained in the \suppref{}). The results for FPN models trained and evaluated on \datalvis{} are shown in Tab.~\ref{tab:zero_shot_inst_det} (see \suppref{} for cascade model results). We notice an interesting trend. For all models, L1 $ > $ L3 $ > $ L2 . This is due to the trade-off between two opposing factors: (a) more complex queries provide more information about the object instance, making L3 task easier than L2, but (b) complex queries also cause errors from multiple attribute predictions to be compounded making L1 better than L3. We include ablation studies in \suppref{} measuring importance of different object and part attributes.


{\noindent\textbf{Comparison with open vocabulary detectors}}\\
To get a sense of the gap between open vocabulary detectors and our task-specific models, we evaluate the publicly available models from Detic~\cite{zhou2022detecting} and MDETR~\cite{kamath2021mdetr} without further fine-tuning on \datalvis{} and report results in Tab.~\ref{tab:zero_shot_open_world} (details in the \suppref{}). In theory, such models can handle arbitrary natural language queries describing object instances. We show results only for L1 queries and two additional subsets: L1 queries with only object attributes ($L1_{obj}$) and only part attributes ($L1_{part}$). We observe limited performance for the evaluated models. This is not surprising and can be attributed to the following factors. Even in the open vocabulary setting, Detic was trained specifically for nouns with little support for attributes. While MDETR was trained for referring expression tasks with attributes, its ability to handle negative images is limited. This highlights the opportunity for future research in open world detectors to handle more descriptive object queries besides category labels.

{\noindent\textbf{Comparison with few-shot models on \dataego{}}}\\
\begin{figure}
\centering
\includegraphics[width=0.9\columnwidth]{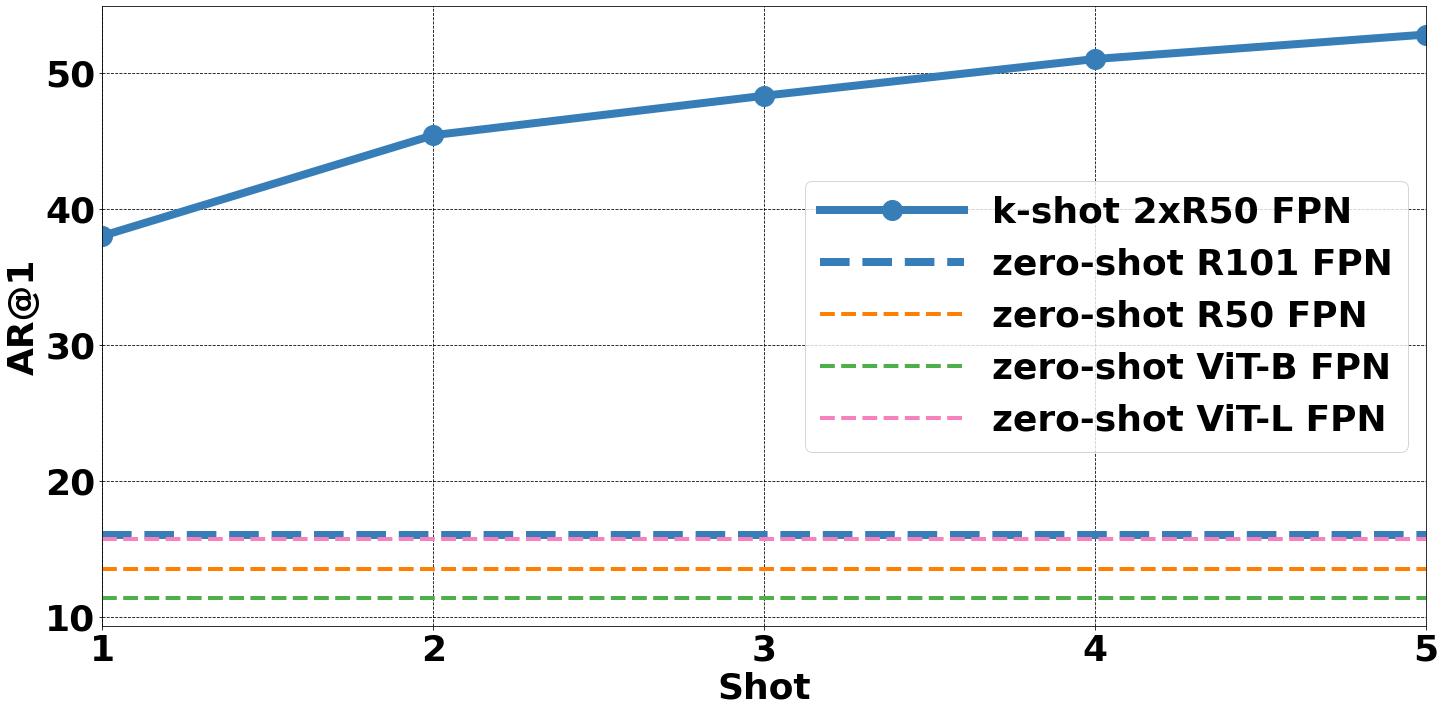}
\caption{Comparing performance of few-shot model with our zero-shot models on \dataego{} instance detection task. Even at $1$-shot we observe a huge gap compared to all zero-shot models.}
\label{fig:k_shot_perf}
\end{figure}
\dataego{} has multiple frames corresponding to the same object instance. Hence, it can serve as a useful dataset for few-shot instance detection as well. Few-shot instance detection is the task where an algorithm is given as input $k$ positive frames with bounding boxes for an object instance and is expected to retrieve another bounding box of the same instance from an unseen set of images. This is similar to our zero-shot task, but the model receives sample object boxes instead of a part/attribute query. We compute and compare zero-shot and few-shot numbers on a subset of $1992$ queries in \dataego{} that have $6$ or more boxes for the object instance corresponding to the query. We benchmark a naive 2-stage model: a pre-trained R50 FPN detector followed by ROI-pooling features from a pre-trained R50 FPN backbone for nearest neighbor ranking (explained in \suppref{}). We evaluate this model for $k$ ranging from $1$-$5$ and compare it to our zero-shot instance detection models trained on the joint \dataname{} dataset in Fig.~\ref{fig:k_shot_perf}. We notice a 20+ point gap even between our best zero-shot model (R101  FPN) and one-shot model ($k=1$). As $k$ increases, the gap widens even further. This shows scope for future improvements to zero-shot object instance detection.



%% file: appendix.tex
\appendix
\section*{Appendix}

\section{Dataset construction}
\label{apx:dataset_const}


\subsection{Parts vocabulary selection}
\begin{figure}
\centering
\includegraphics[width=0.9\columnwidth]{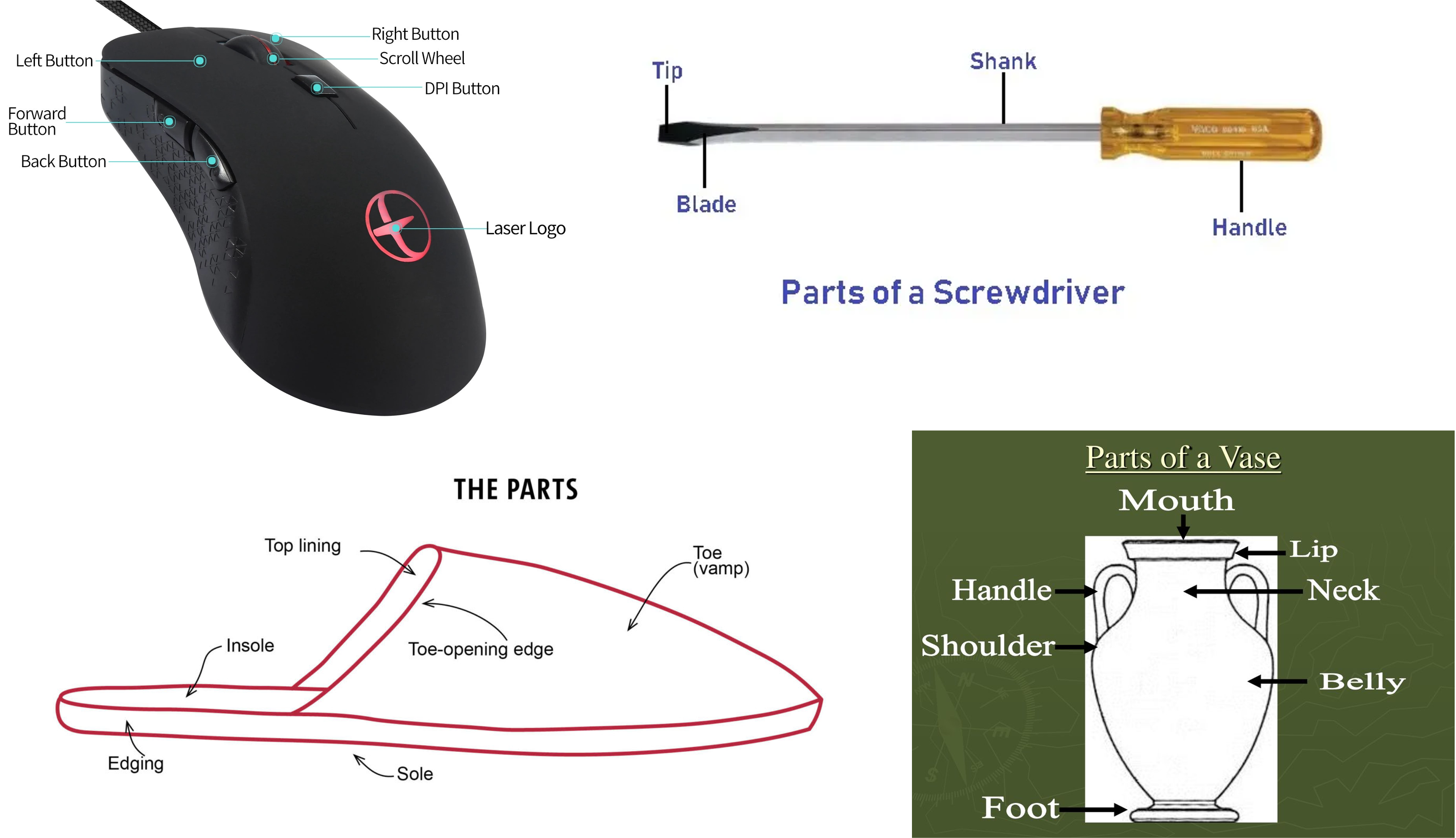}
\caption{Sample web images used to mine part vocabulary. top-left: ``Parts of a computer mouse", top-right: ``Parts of a screwdriver", bottom-left: ``Parts of a slipper" and bottom-right: ``Parts of a vase".}
\label{fig:web_parts}
\end{figure}

\begin{figure}
\centering
\includegraphics[width=0.9\columnwidth]{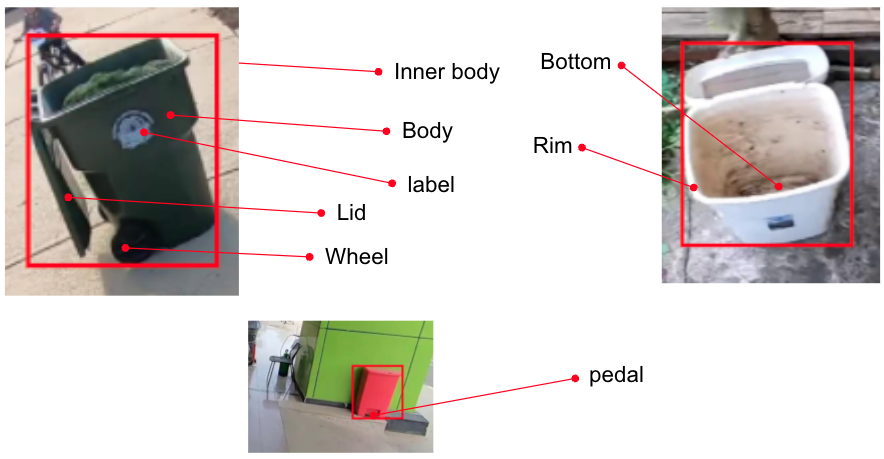}
\caption{Example object category with manually defined parts. For the object ``trash can", we manually defined all the parts with illustrative reference for the annotators as shown.}
\label{fig:manual_parts}
\end{figure}

We show sample images obtained by querying the web for ``parts of an object category" for different object categories in Fig.~\ref{fig:web_parts}. We see that the images provide a good vocabulary of parts for each of the objects. Alongside, they also provide clear pointers to the regions of the object the parts correspond to. We use these as reference images for annotators wherever such well defined part images are available from the web. Additionally, we also manually define parts for few objects when the web images aren't illustrative enough. In such cases, we came up with reasonable names for different regions of an object along with reference images to guide the annotators. Such manually defined parts with sample reference images are shown in Fig.~\ref{fig:manual_parts} as well. Tab.~\ref{tab:part_taxonomy} contains the final taxonomy of parts for the 75 object classes.

\begin{table*}[t]
\vspace*{-0.4in}
    \centering
    \scriptsize
    \begin{tabulary}{0.95\textwidth}{p{.15\textwidth}p{.80\textwidth}}
    \toprule
                         Objects &                                                                                                                                                                                                                Parts Taxonomy \\
    \midrule
                          basket &                                                                                                                                                                        bottom, handle, inner\_side, cover, side, rim, base \\
                            belt &                                                                                                                                                                     buckle, end\_tip, strap, frame, bar, prong, loop, hole \\
                           bench &                                                                                                                                                                                stretcher, seat, back, table\_top, leg, arm \\
                         bicycle &                                                                                                           stem, fork, top\_tube, wheel, basket, seat\_stay, saddle, handlebar, pedal, gear, head\_tube, down\_tube, seat\_tube \\
                         blender &                                                                                                                             cable, handle, cover, spout, vapour\_cover, base, inner\_body, seal\_ring, cup, switch, food\_cup \\
                            book &                                                                                                                                                                                                               page, cover \\
                          bottle &                                                                                                neck, label, shoulder, body, cap, bottom, inner\_body, closure, heel, top, handle, ring, sipper, capsule, spout, base, punt \\
                            bowl &                                                                                                                                                                                       inner\_body, bottom, body, rim, base \\
                             box &                                                                                                                                                                                             bottom, lid, inner\_side, side \\
                           broom &                                                                                                                                                                     lower\_bristles, handle, brush\_cap, ring, shaft, brush \\
                          bucket &                                                                                                                                                                  handle, cover, body, base, inner\_body, bottom, loop, rim \\
                      calculator &                                                                                                                                                                                                                 key, body \\
                             can &                                                                                                                                                                  pull\_tab, body, base, inner\_body, bottom, lid, text, rim \\
                car\_(automobile) & headlight, turnsignal, tank, windshield, mirror, sign, wiper, fender, trunk, windowpane, seat, logo, grille, antenna, hood, splashboard, bumper, rim, handle, runningboard, window, roof, wheel, taillight, steeringwheel \\
                          carton &                                                                                                                                                               inner\_side, tapering\_top, cap, bottom, lid, text, side, top \\
              cellular\_telephone &                                                                                                                                                                                         button, screen, bezel, back\_cover \\
                           chair &                                                                                                                                  stretcher, swivel, apron, wheel, leg, base, spindle, seat, back, rail, stile, skirt, arm \\
                           clock &                                                                                                                                                                     cable, decoration, hand, pediment, finial, case, base \\
                           crate &                                                                                                                                                                                     bottom, handle, inner\_side, lid, side \\
                             cup &                                                                                                                                                                                             inner\_body, handle, rim, base \\
                             dog &                                                                                                                                                                  teeth, neck, foot, head, body, nose, leg, tail, ear, eye \\
                           drill &                                                                                                                                                                                                              handle, body \\
       drum\_(musical\_instrument) &                                                                                                                                                                                   head, rim, cover, body, loop, lug, base \\
                        earphone &                                                                                                                                                                                headband, cable, ear\_pads, housing, slider \\
                             fan &                                                                                                                                         rod, canopy, motor, blade, base, string, light, bracket, fan\_box, pedestal\_column \\
         glass\_(drink\_container) &                                                                                                                                                                                       inner\_body, bottom, body, rim, base \\
                          guitar &                                                                                                                                            key, headstock, bridge, body, fingerboard, back, string, side, pickguard, hole \\
                          hammer &                                                                                                                                                                                                  handle, face, head, grip \\
                         handbag &                                                                                                                                                                          zip, inner\_body, handle, bottom, body, rim, base \\
                             hat &                                                                                                                                                                              logo, pom\_pom, inner\_side, strap, visor, rim \\
                          helmet &                                                                                                                                                                          face\_shield, logo, inner\_side, strap, visor, rim \\
                             jar &                                                                                                                                                           handle, body, base, inner\_body, bottom, lid, sticker, text, rim \\
                          kettle &                                                                                                                                                                                     cable, handle, lid, body, spout, base \\
                           knife &                                                                                                                                                                                                             handle, blade \\
                          ladder &                                                                                                                                                                                                 rail, step, top\_cap, foot \\
                            lamp &                                                                                                                                               shade\_inner\_side, cable, pipe, shade, bulb, shade\_cap, base, switch, finial \\
                 laptop\_computer &                                                                                                                                                         cable, camera, base\_panel, keyboard, logo, back, screen, touchpad \\
                  microwave\_oven &                                                                                                                                          inner\_side, door\_handle, time\_display, control\_panel, turntable, dial, side, top \\
                          mirror &                                                                                                                                                                                                                     frame \\
      mouse\_(computer\_equipment) &                                                                                                                                                    logo, scroll\_wheel, body, right\_button, wire, side\_button, left\_button \\
                             mug &                                                                                                                                                                handle, body, base, inner\_body, bottom, text, drawing, rim \\
                       newspaper &                                                                                                                                                                                                                      text \\
               pan\_(for\_cooking) &                                                                                                                                                                          bottom, handle, inner\_side, lid, side, rim, base \\
                             pen &                                                                                                                                                                                              cap, grip, barrel, clip, tip \\
                          pencil &                                                                                                                                                                                               body, lead, eraser, ferrule \\
                          pillow &                                                                                                                                                                                                                embroidery \\
                            pipe &                                                                                                                                                                                          nozzle, colied\_tube, nozzle\_stem \\
                     plastic\_bag &                                                                                                                                                                                       inner\_body, handle, text, hem, body \\
                           plate &                                                                                                                                                                                  top, bottom, inner\_wall, body, rim, base \\
                          pliers &                                                                                                                                                                                                 jaw, handle, joint, blade \\
                  remote\_control &                                                                                                                                                                                                        logo, back, button \\
                           scarf &                                                                                                                                                                                                             fringes, body \\
                        scissors &                                                                                                                                                                                         handle, screw, finger\_hole, blade \\
                     screwdriver &                                                                                                                                                                                                 blade, handle, tip, shank \\
                            shoe &                                                                                                                       toe\_box, tongue, vamp, outsole, insole, backstay, lining, quarter, heel, throat, eyelet, lace, welt \\
              slipper\_(footwear) &                                                                                                                                                                             toe\_box, vamp, outsole, strap, insole, lining \\
                            soap &                                                                                                                 neck, label, shoulder, body, sipper, capsule, spout, push\_pull\_cap, cap, base, bottom, closure, punt, top \\
                          sponge &                                                                                                                                                                                                             rough\_surface \\
                           spoon &                                                                                                                                                                                                   neck, handle, bowl, tip \\
                           stool &                                                                                                                                                                                                 seat, leg, step, footrest \\
                         sweater &                                                                                                                                                                         shoulder, sleeve, neckband, hem, body, yoke, cuff \\
                           table &                                                                                                                                                         stretcher, drawer, inner\_wall, shelf, apron, wheel, leg, top, rim \\
    tape\_(sticky\_cloth\_or\_paper) &                                                                                                                                                                                                                      roll \\
                       telephone &                                                                                                                                                                                         button, screen, bezel, back\_cover \\
                  television\_set &                                                                                                                                                                                           bottom, button, side, top, base \\
                    tissue\_paper &                                                                                                                                                                                                                      roll \\
                           towel &                                                                                                                                                                                              body, terry\_bar, hem, border \\
                       trash\_can &                                                                                                                                                             label, body, wheel, inner\_body, bottom, lid, pedal, rim, hole \\
                            tray &                                                                                                                                                                                 bottom, inner\_side, outer\_side, rim, base \\
                            vase &                                                                                                                                                                                           neck, handle, foot, body, mouth \\
                          wallet &                                                                                                                                                                                                          inner\_body, flap \\
                          watch &   buckle, case, dial, hand, strap, window, lug \\
                          wrench &                                                                                                                                                                                                              handle, head \\
    \bottomrule
    \end{tabulary}
    \caption{Parts taxonomy}
    \label{tab:part_taxonomy}
\end{table*}

\begin{table*}
  \scriptsize
  \begin{tabulary}{0.95\textwidth}{p{0.1\textwidth}p{0.85\textwidth}}
  \toprule
  Attribute Type & Attribute Classes \\
  \midrule
    Color & black, light\_blue, blue, dark\_blue, light\_brown, brown, dark\_brown, light\_green, green, dark\_green, light\_grey, grey, dark\_grey, light\_orange, orange, dark\_orange, light\_pink, pink, dark\_pink, light\_purple, purple, dark\_purple, light\_red, red, dark\_red, white, light\_yellow, yellow, dark\_yellow \\
    Pattern-Markings & plain, striped, dotted, checkered, woven, studded, perforated, floral, logo, text \\
    Material & stone, wood, rattan, fabric, crochet, wool, leather, velvet, metal, paper, plastic, glass, ceramic \\
    Reflectance & opaque, translucent, transparent \\
    \bottomrule
    \end{tabulary}
    \caption{Attributes taxonomy}
    \label{tab:attr_taxonomy}
\end{table*}

\subsection{Attribute vocabulary selection}

For zero-shot instance recognition tasks, both object and part level attributes are important. In order to identify the set of attributes that we should annotate that are sufficient for the tasks, we conducted an in-depth user study. 

We consider the following $5$ attributes types: color, shape, reflectance, materials and patterns \& marking with the aim of finding a small subset that is sufficient to discriminate between the instances. We show each user two different instances A and B of the same object (green mug and red mug for example), segmentation mask of common object parts between this pair. We use PACO-Ego4D data for this purpose. For object level attributes, we ask annotators to provide at most one difference (if any) for each attribute type. For part level attributes, annotators are asked to compare only between the common parts of the instance pair and they are allowed to annotate up to 3 part level attribute differences for one pair. For an attribute difference, if the discriminative attributes for A and B is nameable (e.g. A is red and B is blue), annotators will need to write down the attribute names. Otherwise, a freeform explanation is required to articulate this difference, particularly for unnameable attributes (e.g. a unique pattern, an irregular shape, etc.);
For each object category we sampled $106$ pairs each which are annotated by $3$ annotators. 

{\bf{Un-nameable shape attributes.}} Annotators noted that $>50\%$ shape differences contain unnameable attributes. Annotators reported these differences as very difficult to describe with words. Hence, we removed ``shape" from the final list of attribute. Nevertheless, even in the absence of shape we note that the combination of the remaining attributes are seen to be sufficiently discriminative to differentiate the object instances.

{\bf{Attributes Coverage.}} We try to identify the discriminative power of different subsets of attributes and identify the best subset to construct our attributes taxonomy. We adopted a greedy algorithm to study attribute sets. We start with one attribute and gradually add one best attribute at a time to incrementally construct an attribute set at each step. More specifically at a give step, for each attribute, we check how many new pairs can be distinguished if we introduce that attribute to the existing set of attributes. The attribute that distinguishes highest number of pairs is selected first, followed by the next best attribute in a greedy fashion. We define coverage of a set of attributes as the total number of object pairs that can be distinguished by the attributes (both with object-level attributes and/or part-level attributes).

We observed that coverage plateaus at $40$ attributes. $98\%$ of object instance pairs could be distinguished only using the $55$ attributes included in our final version of PACO. Both object and part attributes were marked as important for differentiating instance pairs. $~18\%$ instance pairs could only be distinguished by object level attributes, while $~10\%$ could only be distinguished by part level attributes. Color is the biggest discriminative attribute type for instance recognition, differentiating at least $~75\%$ instance pairs with both object and part level color differences.

The final taxonomy of attributes is shown in Tab.~\ref{tab:attr_taxonomy}.

\subsection{Annotation pipeline}

\subsubsection{Instance annotation}
To enable appearance based k-shot instance detection experiments we have annotated instances with unique instance IDs. For LVIS (image) dataset, we assume image of each object to be a separate instance. We inspected several images manually and found this assumption to be true. In Ego4D videos from which we sourced the frames, however, the same instance can occur multiple times at different timestamps and we had to set up an annotation task to properly group occurrences (frames) into instances. There are two challenges that we faced: (a) the same video can contain different instances of the same object class and those have to be split into separate instance IDs, and (b) Ego4D videos are fragmented and multiple videos can contain the same instances so occurrences from different videos had to be merged. To this end we performed a three-step splitting/merging annotation pipeline as follows. \\

{\noindent{\textbf{Split:}}} Using (video, category) pair as a good first guess for  instance ID, we crop all the bounding boxes (occurrences) of an object category from frames that belong to the same video and show them to annotators. We then ask the annotators to split those crops into subgroups that belong to the same real instance. In case number of boxes is more than $16$, we split them in to groups of at most $16$ and then send them for annotation. This is then repeated for all object categories and all videos. All annotation jobs are reviewed by 3 annotators and a subset majority voting is performed to aggregate annotations. The majority voting is done by finding the maximum overlap between subgroups for each pair of annotators using Hungarian algorithm (bipartite matching). \\

{\noindent{\textbf{Merge:}}} After the splitting phase the annotated groups are very coherent, i.e., the majority of occurrences in the same group belong to the same instance. However due to video fragmentation and additional limitation on the number of boxes that can be shown to annotators ($16$) many occurrence groups belong to the same instance and need to be merged. To address this we use similarity in DINO model~\cite{caron2021emerging} embedding space. Each group from the splitting phase is represented by a bounding box crop with embedding closest to the group median. For each group representative $g_i$ we find $16$ nearest neighbors and ask annotators to validate which of the neighbors belong to the same instance as $g_i$. Similar to the splitting phase, responses from 3 annotators are aggregated by finding the maximum overlap between any two annotators. We repeat this for every group. We then build a graph by considering each group as a node with an edge between two nodes if they belong to the same instance. Nodes $i$ and $j$ are connected if $g_j$ was marked as belonging to the same instance as $g_i$ \textbf{\emph{and}} $g_i$ was marked as belonging to the same instance as $g_j$. Finally, we find connected components and assign a unique instance ID to each component.  \\

{\noindent{\textbf{Final split:}}} We noticed some over-merging of instances, especially for instances with large number of occurrences. We therefore performed a third step where we showed instances with more than $10$ occurrences to expert annotators and asked them to split them into subgroups. Each subgroup at the output of this step is then marked as a separate instance. There was no limit of $16$ occurrences in this step, complete instances were shown in each annotation job.

\subsubsection{Managing annotation quality}

\begin{figure}
    \centering
    \includegraphics[width=0.6\linewidth]{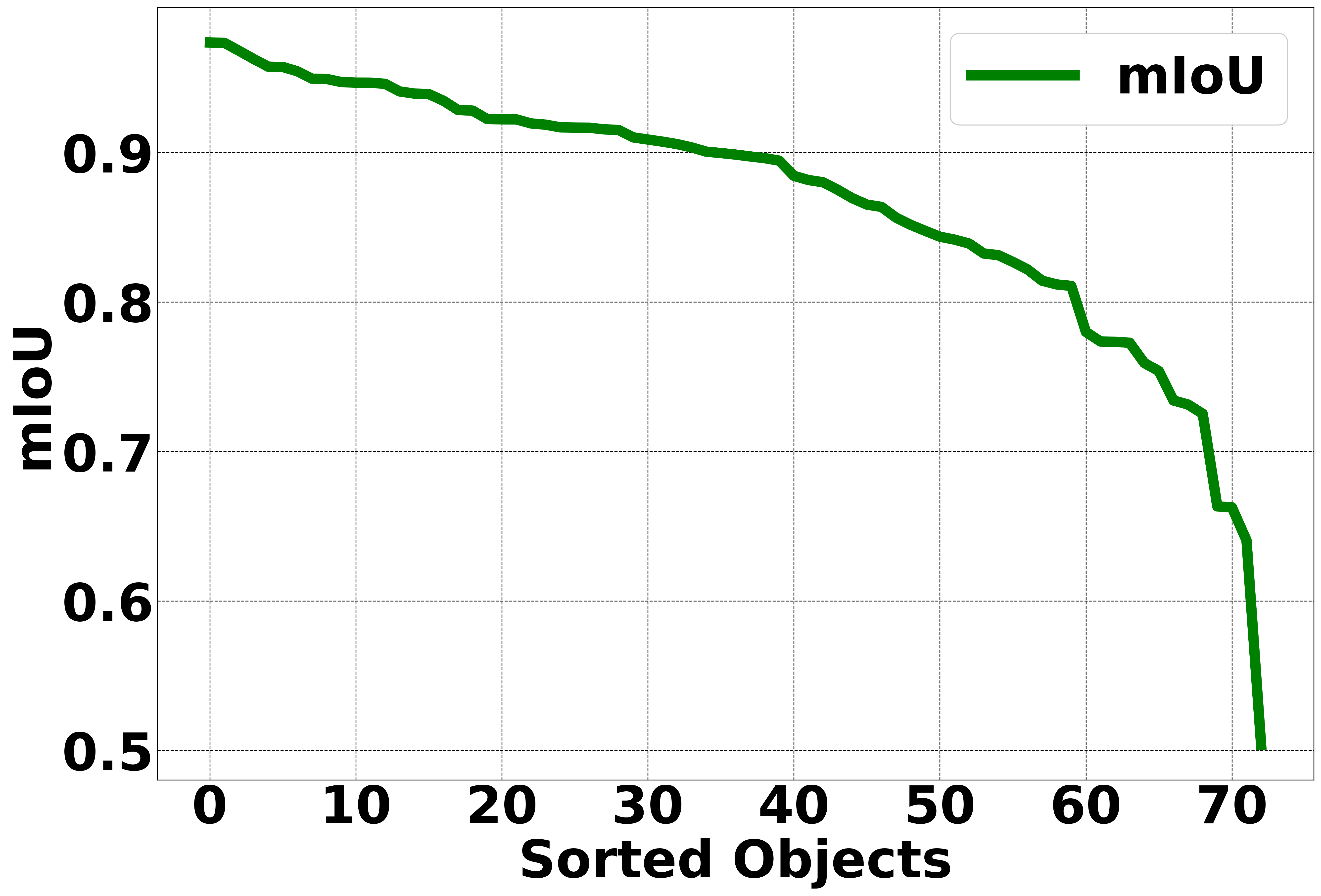}
    \caption{Distribution of mIoU with gold-standard part masks for different object classes. $90\%$ of the object classes have mIoU $\ge 0.75$ with the gold-standard masks.}
    \label{fig:gold_miou}
\end{figure}

Fig.~\ref{fig:gold_miou} shows the mIoU of annotated masks with gold set masks for each object category.

\begin{figure*}
\centering
\includegraphics[width=0.98\textwidth]{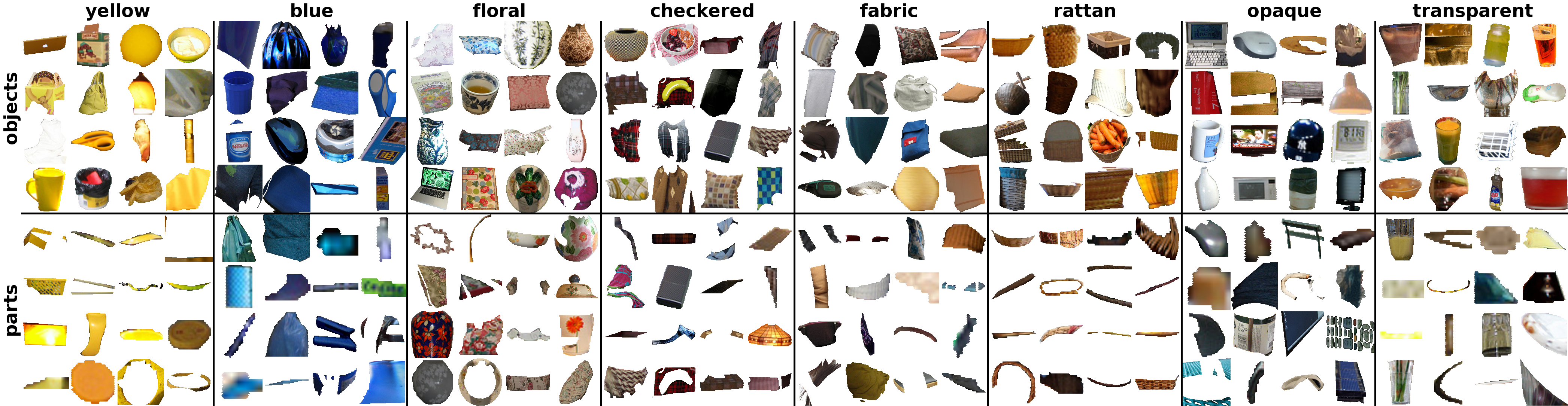}
\caption{Randomly sampled object (top row) and part (bottom row) masks for a subset of attributes (one attribute per column).}
\label{fig:attr_collage}
\end{figure*}

\section{Dataset annotation examples}
\label{apx:dataset_examples}
\begin{figure*}
\centering
\includegraphics[width=0.95\textwidth]{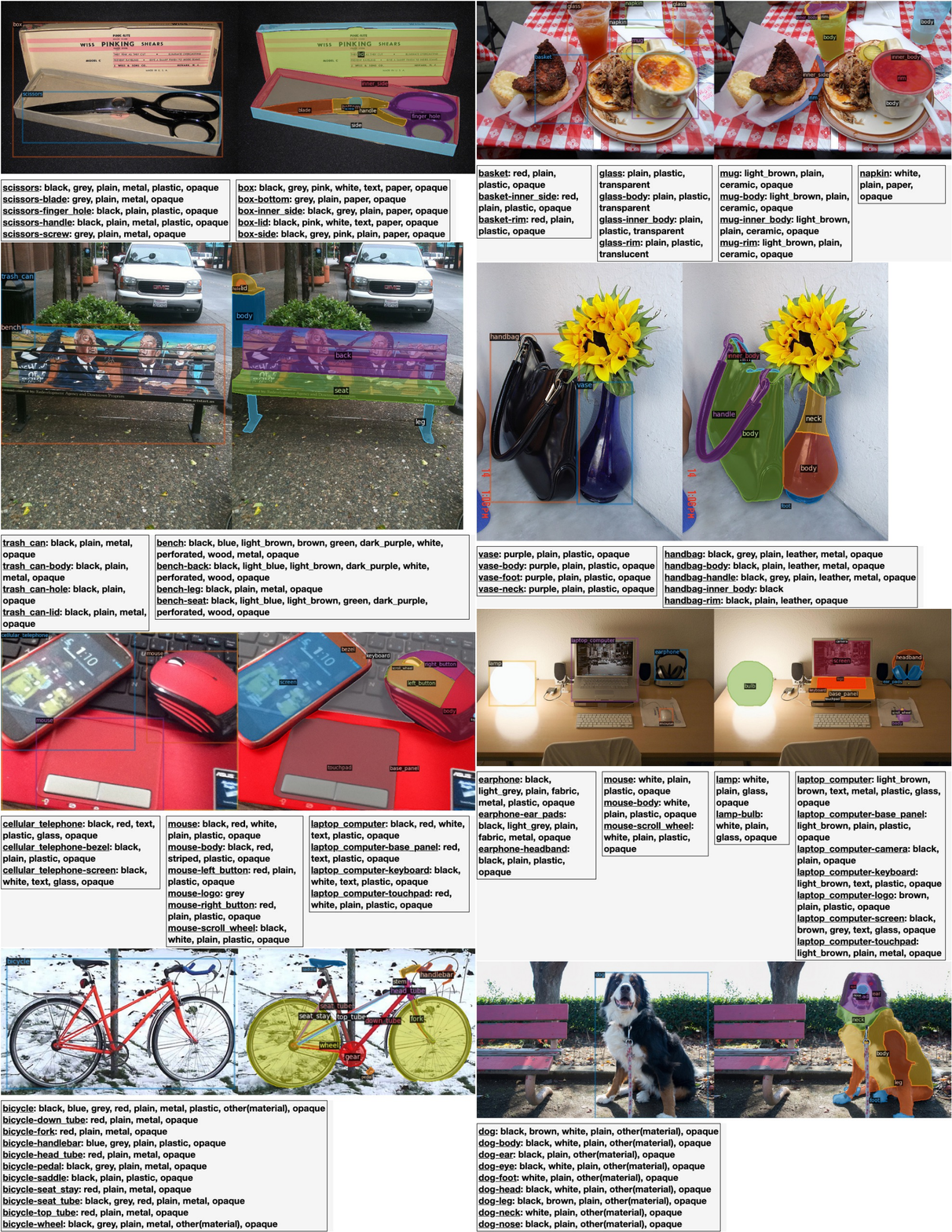}
\vspace*{-1.5pt}
\caption{Annotation examples. Each image contains object bounding boxes (object masks omitted so attributes are visible) on the left and part masks on the right. Object and part attributes are listed below each image.}
\label{fig:ann_examples}
\end{figure*}

Object, part, and attribute annotations are shown in Figs.~\ref{fig:attr_collage} and \ref{fig:ann_examples}. Object and part segmentation masks are used to crop out segments for annotations with a specific attribute and shown in Fig.~\ref{fig:attr_collage} for a subset of attributes. Fig.~\ref{fig:ann_examples} shows various examples for \dataname{} annotations. Full images are shown with object annotations (bounding boxes only so attributes are visible) in the left copy of the image and part annotations (segmentation masks) in the right copy of the image. Object and part attribute annotations are listed below each image pair.

\section{Object statistics}
\label{apx:dataset_stats}

\begin{figure}
    \centering
    \includegraphics[width=0.6\linewidth]{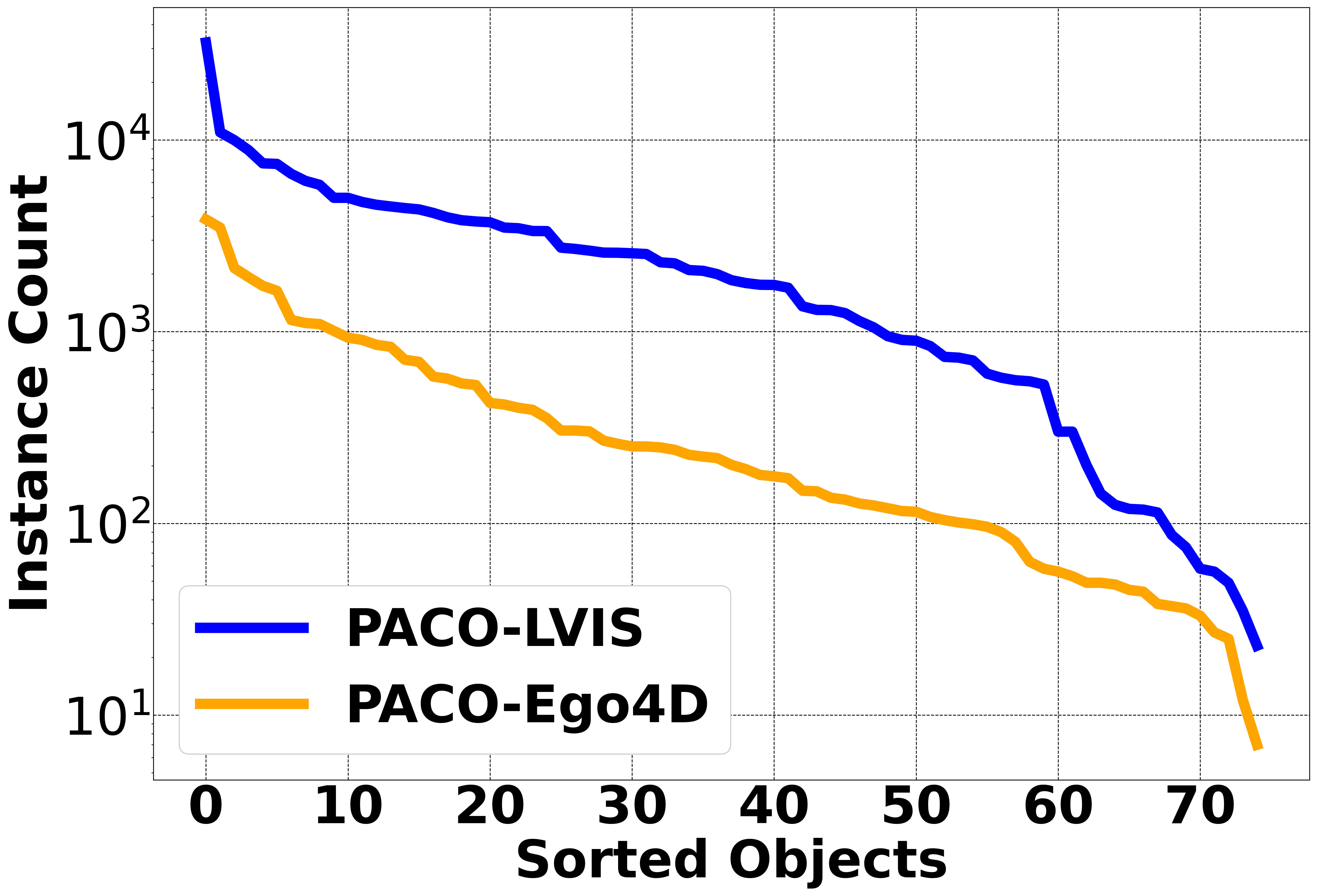}
    \caption{Distribution of instances across the $75$ object categories.}
    \label{fig:obj stats}
\end{figure}

Fig.~\ref{fig:obj stats} shows the distribution of instances across the $75$ object categories in \datalvis{} and \dataego{}. All $75$ object classes in \datalvis{} and $71$ classes in \dataego{} have $\ge 10$ instances. We observe the usual non-uniformity in the frequency for each category. For object category `drill' with the lowest frequency in  \datalvis{}, we have $23$ instances, and for `scarf' with the lowest frequency in \dataego{} data, we have $7$ instances.

\section{Additional part segmentation and attribute prediction results}
\label{apx:part_attr_res}
\begin{figure*}[t]
    \centering
    \includegraphics[width=0.8\textwidth]{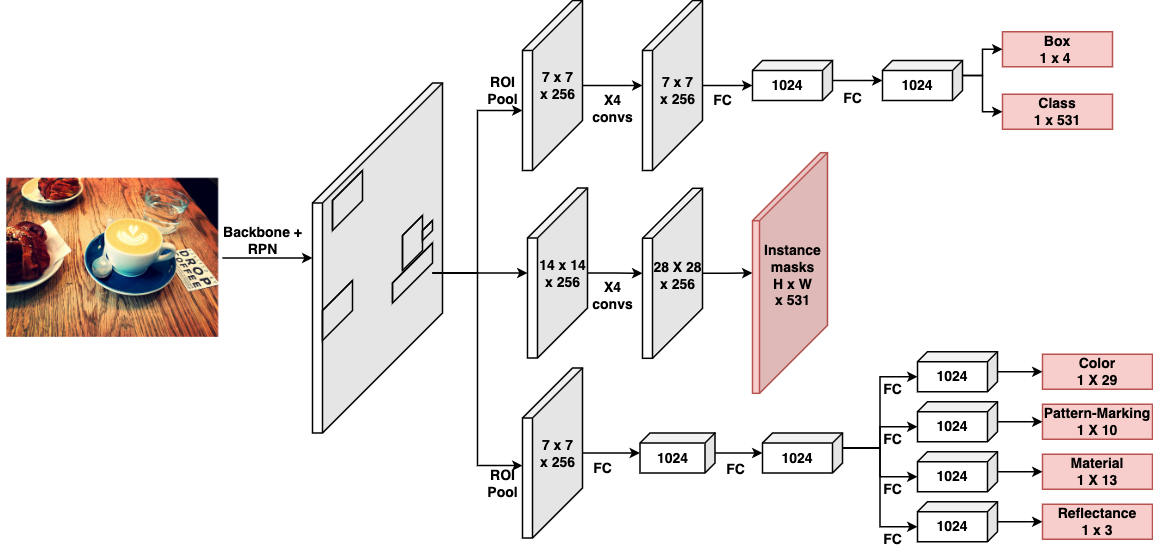}
    \caption{Our model adds an attribute prediction head to
Mask R-CNN for joint instance segmentation with attribute prediction}
    \label{fig:model_arch}
\end{figure*}

In Fig.~\ref{fig:model_arch}, we show the architecture of the models used to train the joint segmentation and attribute prediction models. For our experiments, we vary the backbones across R-50, R-101 and two ViT-det~\cite{Li2022ExploringPV} model backbones.

\begin{figure*}
\centering
\includegraphics[width=0.95\textwidth]{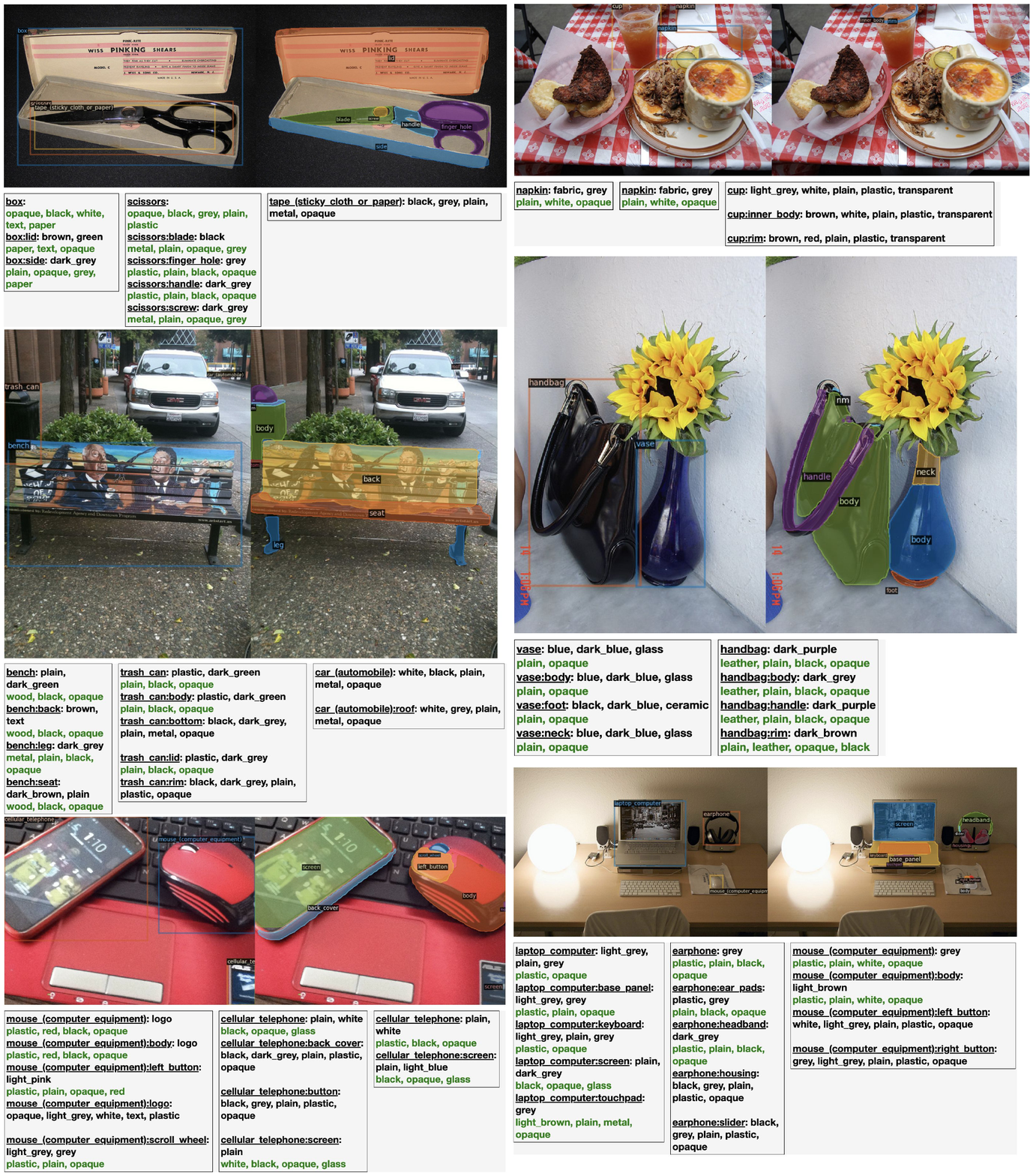}
\vspace*{-1.0in}
\caption{Part segmentation and attribute prediction examples from a Vit-L model trained on \datalvis{} and \dataego{}. Each image contains predicted object bounding boxes for the 3 highest scoring objects on the left and predicted part masks which overlap with these objects on the right. The corresponding object and part attribute predictions are listed below each image. Attribute predictions in green are contained in ground truth.}
\label{fig:predictions_examples}
\end{figure*}

Examples of predictions from the ViT-L model are shown in Fig.~\ref{fig:predictions_examples}.

\begin{table}
    \setlength{\tabcolsep}{2mm}  
    \centering
    \footnotesize
    \begin{tabular}{c||c c|c c}
    \hline
    & \multicolumn{2}{c|}{mask $AP$} & \multicolumn{2}{c}{box $AP$} \\
    Model & \apobj & \apopart & \apobj & \apopart\\
    \hline \hline
    R50 FPN & 31.2 \deemph{$\pm$ 0.1} & 12.1 \deemph{$\pm$ 0.1} & 34.3 \deemph{$\pm$ 0.2} & 15.7 \deemph{$\pm$ 0.2} \\
    \hline
    R101 FPN & 32.0 \deemph{$\pm$ 0.3} & 12.5 \deemph{$\pm$ 0.1} & 35.2 \deemph{$\pm$ 0.3} & 16.2 \deemph{$\pm$ 0.2} \\
    \hline
    ViT-B FPN & 35.5 \deemph{$\pm$ 0.5} & 14.1 \deemph{$\pm$ 0.3} & 39.2 \deemph{$\pm$ 0.5} & 18.1 \deemph{$\pm$ 0.5} \\
    \hline
    ViT-L FPN & 44.7 \deemph{$\pm$ 0.4} & 18.1 \deemph{$\pm$ 0.3} & 49.6 \deemph{$\pm$ 0.4} & 22.9 \deemph{$\pm$ 0.4} \\
    \hline
    \end{tabular}
    \caption{Object and object-part segmentation results for mask-RCNN and ViT-det models trained jointly on PACO-LVIS and PACO-EGO4D and evaluated on PACO-LVIS}
    \label{tab:part_seg_results_lvis}
\end{table}

\begin{table}
    \setlength{\tabcolsep}{2mm}  
    \centering
    \footnotesize
    \begin{tabular}{c||c c|c c}
    \hline
    & \multicolumn{2}{c|}{mask $AP$} & \multicolumn{2}{c}{box $AP$} \\
    Model & \apobj & \apopart & \apobj & \apopart\\
    \hline \hline
    R50 FPN & 16.6 \deemph{$\pm$ 0.3} & 5.6 \deemph{$\pm$ 0.0} & 18.9 \deemph{$\pm$ 0.3} & 8.2 \deemph{$\pm$ 0.1} \\
    \hline
    R101 FPN & 17.9 \deemph{$\pm$ 0.2} & 6.0 \deemph{$\pm$ 0.1} & 20.3 \deemph{$\pm$ 0.2} & 8.7 \deemph{$\pm$ 0.1} \\
    \hline
    ViT-B FPN & 18.6 \deemph{$\pm$ 0.2} & 7.0 \deemph{$\pm$ 0.2} & 20.7 \deemph{$\pm$ 0.3} & 10.1 \deemph{$\pm$ 0.1} \\
    \hline
    ViT-L FPN & 27.9 \deemph{$\pm$ 0.3} & 10.5 \deemph{$\pm$ 0.3} & 30.6 \deemph{$\pm$ 0.2} & 14.8 \deemph{$\pm$ 0.4} \\
    \hline
    \end{tabular}
    \caption{Object and object-part segmentation results for mask-RCNN and ViT-det models trained jointly on PACO-LVIS and PACO-EGO4D and evaluated on PACO-EGO4D}
    \label{tab:part_seg_results_ego4d}
\end{table}

\begin{table*}
    \setlength{\tabcolsep}{1.5mm}  
    \centering
    \footnotesize
    \begin{tabular}{c || c | c c c c || c | c c c c }
    \hline
    Model & $AP^{obj}_{att}$ & $AP^{obj}_{col}$ & $AP^{obj}_{pat}$ & $AP^{obj}_{mat}$ & $AP^{obj}_{ref}$ & $AP^{opart}_{att}$ & $AP^{opart}_{col}$ & $AP^{opart}_{pat}$ & $AP^{opart}_{mat}$ & $AP^{opart}_{ref}$ \\
    \hline \hline
    R50 FPN & 13.8 \deemph{$\pm$ 0.1} & 10.6 \deemph{$\pm$ 0.4} & 14.9 \deemph{$\pm$ 0.7} & 9.7 \deemph{$\pm$ 0.2} & 19.8 \deemph{$\pm$ 0.9} & 9.7 \deemph{$\pm$ 0.1} & 10.3 \deemph{$\pm$ 0.5} & 10.7 \deemph{$\pm$ 0.5} & 7.2 \deemph{$\pm$ 0.2} & 10.7 \deemph{$\pm$ 0.2} \\
    \hline
      R101 FPN & 14.0 \deemph{$\pm$ 0.4} & 11.2 \deemph{$\pm$ 0.3} & 14.2 \deemph{$\pm$ 0.9} & 9.8 \deemph{$\pm$ 0.4} & 20.6 \deemph{$\pm$ 1.6} & 10.1 \deemph{$\pm$ 0.2} & 10.8 \deemph{$\pm$ 0.4} & 11.0 \deemph{$\pm$ 0.3} & 7.2 \deemph{$\pm$ 0.0} & 11.3 \deemph{$\pm$ 0.3} \\
    \hline
      ViT-B FPN & 16.2 \deemph{$\pm$ 0.6} & 13.2 \deemph{$\pm$ 0.4} & 16.7 \deemph{$\pm$ 0.9} & 13.3 \deemph{$\pm$ 0.3} & 21.4 \deemph{$\pm$ 1.4} & 11.5 \deemph{$\pm$ 0.1} & 12.0 \deemph{$\pm$ 0.1} & 12.6 \deemph{$\pm$ 0.2} & 9.4 \deemph{$\pm$ 0.0} & 11.8 \deemph{$\pm$ 0.4} \\
    \hline
      ViT-L FPN & 18.8 \deemph{$\pm$ 0.7} & 15.6 \deemph{$\pm$ 0.2} & 19.6 \deemph{$\pm$ 1.1} & 15.7 \deemph{$\pm$ 0.6} & 24.5 \deemph{$\pm$ 1.2} & 14.1 \deemph{$\pm$ 0.1} & 15.0 \deemph{$\pm$ 0.3} & 15.2 \deemph{$\pm$ 0.7} & 11.6 \deemph{$\pm$ 0.1} & 14.3 \deemph{$\pm$ 0.2} \\
    \end{tabular}
    \caption{Attribute prediction results for a mask R-CNN and ViT-det model trained jointly on \datalvis{} and \dataego{} and evaluated on \datalvis{}. The results are shown for box $AP$ for both object attributes and object-part attributes prediction.}
    \label{tab:attr_results_lvis}
\end{table*}

\begin{table*}
    \setlength{\tabcolsep}{1.5mm}  
    \centering
    \footnotesize
    \begin{tabular}{c || c | c c c c || c | c c c c }
    \hline
    Model & $AP^{obj}_{att}$ & $AP^{obj}_{col}$ & $AP^{obj}_{pat}$ & $AP^{obj}_{mat}$ & $AP^{obj}_{ref}$ & $AP^{opart}_{att}$ & $AP^{opart}_{col}$ & $AP^{opart}_{pat}$ & $AP^{opart}_{mat}$ & $AP^{opart}_{ref}$ \\
    \hline \hline
    R50 FPN & 6.6 \deemph{$\pm$ 0.4} & 5.2 \deemph{$\pm$ 0.2} & 7.0 \deemph{$\pm$ 0.3} & 6.6 \deemph{$\pm$ 0.8} & 7.7 \deemph{$\pm$ 0.4} & 5.6 \deemph{$\pm$ 0.1} & 5.6 \deemph{$\pm$ 0.5} & 6.6 \deemph{$\pm$ 0.6} & 5.7 \deemph{$\pm$ 0.3} & 4.5 \deemph{$\pm$ 0.3} \\
    \hline
      R101 FPN & 7.3 \deemph{$\pm$ 0.2} & 5.4 \deemph{$\pm$ 0.2} & 7.6 \deemph{$\pm$ 0.3} & 8.1 \deemph{$\pm$ 0.5} & 8.2 \deemph{$\pm$ 0.4} & 5.9 \deemph{$\pm$ 0.1} & 5.7 \deemph{$\pm$ 0.6} & 7.0 \deemph{$\pm$ 1.1} & 6.1 \deemph{$\pm$ 0.3} & 4.6 \deemph{$\pm$ 0.3} \\
    \hline
      ViT-B FPN & 8.6 \deemph{$\pm$ 0.1} & 6.6 \deemph{$\pm$ 0.5} & 10.8 \deemph{$\pm$ 0.7} & 8.7 \deemph{$\pm$ 0.3} & 8.2 \deemph{$\pm$ 0.7} & 7.3 \deemph{$\pm$ 0.1} & 6.2 \deemph{$\pm$ 0.8} & 10.7 \deemph{$\pm$ 0.4} & 6.8 \deemph{$\pm$ 0.6} & 5.7 \deemph{$\pm$ 0.0} \\
    \hline
      ViT-L FPN & 11.7 \deemph{$\pm$ 0.3} & 9.0 \deemph{$\pm$ 0.1} & 13.1 \deemph{$\pm$ 1.5} & 12.4 \deemph{$\pm$ 0.2} & 12.4 \deemph{$\pm$ 0.4} & 10.0 \deemph{$\pm$ 0.5} & 7.8 \deemph{$\pm$ 0.6} & 12.6 \deemph{$\pm$ 2.0} & 9.9 \deemph{$\pm$ 0.2} & 9.7 \deemph{$\pm$ 0.3} \\
    \end{tabular}
    \caption{Attribute prediction results for a mask R-CNN and ViT-det model trained jointly on \datalvis{} and \dataego{} and evaluated on \dataego{}. The results are shown for box $AP$ for both object attributes and object-part attributes prediction.}
    \label{tab:attr_results_ego4d}
\end{table*}

\begin{table}
    \setlength{\tabcolsep}{1.3mm}  
    \centering
    \footnotesize
    \begin{tabular}{c||c c c c c}
    \hline

    Model & split & \apobj & \apopart & $AP^{obj}_{att}$ & $AP^{opart}_{att}$ \\
    \hline \hline
    R50 FPN & val & 38.3 \deemph{$\pm$ 0.4} & 18.4 \deemph{$\pm$ 0.3} & 20.0 \deemph{$\pm$ 0.5} & 17.1 \deemph{$\pm$ 0.4} \\
    & test & 34.3 \deemph{$\pm$ 0.2} & 15.7 \deemph{$\pm$ 0.2} & 13.8 \deemph{$\pm$ 0.1} & 9.7 \deemph{$\pm$ 0.1} \\

    \hline
    R101 FPN  & val & 39.4 \deemph{$\pm$ 0.2} & 18.7 \deemph{$\pm$ 0.5} & 21.0 \deemph{$\pm$ 0.9} & 17.3 \deemph{$\pm$ 0.4} \\
    & test & 35.2 \deemph{$\pm$ 0.3} & 16.2 \deemph{$\pm$ 0.2} & 14.0 \deemph{$\pm$ 0.4} & 10.1 \deemph{$\pm$ 0.2} \\

    \hline
    ViT-B FPN & val & 42.6 \deemph{$\pm$ 0.7} & 20.8 \deemph{$\pm$ 0.7} & 25.2 \deemph{$\pm$ 0.5} & 21.0 \deemph{$\pm$ 0.3} \\
    & test & 39.2 \deemph{$\pm$ 0.5} & 18.1 \deemph{$\pm$ 0.5} & 16.2 \deemph{$\pm$ 0.6} & 11.5 \deemph{$\pm$ 0.1} \\

    \hline
    ViT-L FPN & val & 52.6 \deemph{$\pm$ 0.5} & 25.9 \deemph{$\pm$ 0.7} & 29.2 \deemph{$\pm$ 0.6} & 25.9 \deemph{$\pm$ 0.1} \\

    & test & 49.6 \deemph{$\pm$ 0.4} & 22.9 \deemph{$\pm$ 0.4} & 18.8 \deemph{$\pm$ 0.7} & 14.1 \deemph{$\pm$ 0.1} \\
    \hline
    \end{tabular}
    \caption{We compare how object detection, object-part detection and attribute prediction results transfer from \texttt{val} set to \texttt{test} set. The models are trained jointly on PACO-LVIS and PACO-EGO4D and evaluated on PACO-LVIS. The ranking is consistent across both splits}
    \label{tab:test_val_trends_lvis}
\end{table}

\begin{table}
    \setlength{\tabcolsep}{1.3mm}  
    \centering
    \footnotesize
    \begin{tabular}{c||c c c c c}
    \hline

    Model & split & \apobj & \apopart & $AP^{obj}_{att}$ & $AP^{opart}_{att}$ \\
    \hline \hline
    R50 FPN & val& 37.3 \deemph{$\pm$ 0.7} & 18.6 \deemph{$\pm$ 0.4} & 28.8 \deemph{$\pm$ 3.3} & 21.2 \deemph{$\pm$ 3.7} \\
    & test & 18.9 \deemph{$\pm$ 0.3} & 8.2 \deemph{$\pm$ 0.1} & 6.6 \deemph{$\pm$ 0.4} & 5.6 \deemph{$\pm$ 0.1}\\

    \hline
    R101 FPN  & val & 38.9 \deemph{$\pm$ 0.3} & 19.4 \deemph{$\pm$ 0.2} & 30.5 \deemph{$\pm$ 3.4} & 22.7 \deemph{$\pm$ 3.0} \\
    & test & 20.3 \deemph{$\pm$ 0.2} & 8.7 \deemph{$\pm$ 0.1} & 7.3 \deemph{$\pm$ 0.2} & 5.9 \deemph{$\pm$ 0.1} \\

    \hline
    ViT-B FPN & val & 48.1 \deemph{$\pm$ 0.3} & 24.9 \deemph{$\pm$ 0.1} & 44.3 \deemph{$\pm$ 2.2} & 35.5 \deemph{$\pm$ 1.0} \\
    & test & 20.7 \deemph{$\pm$ 0.3} & 10.1 \deemph{$\pm$ 0.1} & 8.6 \deemph{$\pm$ 0.1} & 7.3 \deemph{$\pm$ 0.1} \\

    \hline
    ViT-L FPN & val & 56.1 \deemph{$\pm$ 0.1} & 30.8 \deemph{$\pm$ 0.1} & 48.8 \deemph{$\pm$ 3.1} & 39.8 \deemph{$\pm$ 0.7} \\

    & test & 30.6 \deemph{$\pm$ 0.2} & 14.8 \deemph{$\pm$ 0.4} & 11.7 \deemph{$\pm$ 0.3} & 10.0 \deemph{$\pm$ 0.5} \\
    \hline
    \end{tabular}
    \caption{We compare how object detection, object-part detection and attribute prediction results transfer from \texttt{val} set to \texttt{test} set. The models are trained jointly on PACO-LVIS and PACO-EGO4D and evaluated on PACO-EGO4D. The ranking is consistent across both splits}
    \label{tab:test_val_trends_ego4d}
\end{table}

\subsection{Joint training on \datalvis{} and \dataego{}}
In addition to models trained on \datalvis{}, we also train models for part segmentation and attribute prediction jointly trained on both \datalvis{} and \dataego{}. We evaluate the jointly trained models on the test splits for both the datasets and present the results for part segmentation in Tab.~\ref{tab:part_seg_results_lvis} and Tab.~\ref{tab:part_seg_results_ego4d}. Tab.~\ref{tab:attr_results_lvis} and Tab.~\ref{tab:attr_results_ego4d} show the results on attribute prediction. We notice that the results on \dataego{} overall are lower compared to those for \datalvis{}. This is indicative of the challenges in video domain particularly for ego-centric videos. Also, we note that the jointly trained model offers a small improvement compared to model trained only on \datalvis{}, when evaluate on \datalvis{} in Tab.~\ref{tab:attr_results_lvis}. We observed $0.2\%$ gain for R50-FPN and R101-FPN and $0.8\%$ improvement for ViT-B FPN, compared to model trained only with \datalvis{}.

\subsection{\texttt{val} to \texttt{test} results transfer}
Here, we wish to study if observations made from the \texttt{val} split are similar to the \texttt{test} split. This would help us verify if \texttt{val} split can be used for model tuning. In Tab.~\ref{tab:test_val_trends_lvis} and Tab.~\ref{tab:test_val_trends_ego4d}, we observe that the ranking of results is consistent across \texttt{val} and \texttt{test}. Across different architectures the trends are similar. This study is similar to what is reported in LVIS~\cite{gupta2019lvis} for object detection.

\subsection{Object segmentation only models}
In this section, we explore the effect of joint training on multiple tasks (segmentation and attribute prediction) together on object segmentation results. As an ablation, we train models on only the task of object segmentation for two backbones: R-50 and ViT-L. We report our observations in Tab.~\ref{tab:obj_only}. For the smaller R-50 backbone,  the object segmentation performance deteriorates slightly when joint training with multiple tasks. However, surprisingly for the higher capacity ViT-L backbone, object segmentation improves considerably when training on the joint task.

\subsection{Attribute prediction bounds}
In the main paper, we report the bounds on $AP^{obj}_{att}$. The lower bound is calculated by assuming that the score for the object-attribute prediction is the same as the score for the object prediction, i.e., the lower bound performance is the same as if only the detector was used for attribute prediction. The upper bound performance assumes perfect attribute prediction by setting the score for gt attribute to $1.0$ and any false positive attribute predictions to $0.0$ for a given object prediction. Here, object refers to both object and object-parts.

\section{Additional zero-shot instance detection results}
\label{apx:zero_shot_res}
We show results for FPN and cascade models trained and evaluated on \datalvis{} in Tab.~\ref{tab:zero_shot_inst_det_lvis}. Cascade models improve the performance for all but the largest model. In Tab.~\ref{tab:zero_shot_inst_det_joint} we also show the results from models trained on the joint \dataname{} dataset and evaluated on \datalvis{} and \dataego{} test sets. \dataego{} is a more challenging dataset, zero-shot results are in line with attributes prediction results shown in Tab.~\ref{tab:attr_results_ego4d}.

\section{Ablation studies for zero-shot instance detection}
\label{apx:zero_shot_abl}
We also measure the importance of different aspects such as object category, object-part category, object colors, part colors and non-color attributes for this end to end task by incrementally including them over a vanilla detection model in Tab.~\ref{tab:zero_shot_ablation}. As expected, the object-only performance is poor and each additional component improves the instance detection performance.


\section{From model outputs to query scores}
\label{query_score_comp}
For prediction ranking in the zero-shot instance detection task we need query scores for each detected box. However models trained in Sec.~\attrexp{} produce object, part, and attribute scores instead. In this section we provide details of how these scores are used to obtain query scores for each box.

Let $Q$ be a query for an object $o$, with object-level attributes $A$, parts $P$ and part-level attributes $A_p \forall p \in P$. For example, the query ``Black dog with white ear and brown foot" corresponds to $o$ (dog), object-level attributes $A$ (\{``black"\}), parts $P$ (\{``ear", ``foot"\}), part-level attributes $A_p$ ( \{``white"\} for ``ear", \{``brown"\} for ``foot").

Given such a query, we assign a query score to all object boxes in an image. This is a two-step process. In the first step, we associate object-parts detected by our model to the corresponding object boxes in the image. In the second step, we calculate the query score for each object box based on the associated parts.

{\noindent\textbf{Part association.}} Since object-part and object boxes are detected independently by our model, we need to associate part boxes to objects first. For a given object box, consider all part boxes where the part class corresponds to the object class of the object box, e.g., for a ``car" object box, we will only consider predictions for ``car-wheel" and not ``bicycle-wheel". From these, select part boxes where more than $50\%$ of the part mask area is contained within the object mask. Call these part boxes matched parts. The matched parts may contain multiple occurrences of the same object-part class, keep only the one with the highest score. This results in set of matched parts for each object box. For some objects, we may have no matched parts for a specific object-part (eg: we may find no ``car-wheel" matched with a ``car" box).

{\noindent\textbf{Query score.}} 
For a given object box $b$, let the predicted score for the query object category $o$ be given by $o_o$. Similarly, let the predicted object attribute scores be $a_k$ for $k \in A$. Similarly, the part scores of the matched object-parts are given by $p_p$ for $p \in P$. These are the predicted category scores for the matched part box corresponding to each object-part category mentioned in the query. For an object-part category if no part box is matched to $b$, this score is set to $0$. We also have attribute scores for each matched object-part $a_{p,k}$ for $p \in P, k \in A_p$. These scores are again set to $0$ if no part box of the corresponding object-part category is matched with $b$. For the query $Q$, the score is then computed as follows:
\begin{multline*}
s(Q,b) = \left\{ 
    \begin{array}{ll} 
        \text{if\ }|A|>0: & \sqrt{o_o \times \sqrt[|A|]{\prod_{k\in A}a_k}}  \\
        \text{otherwise}: & o_o 
    \end{array}
\right. \\
\times
\left\{
    \begin{array}{ll} 
        \text{if\ }|P|>0: & 
            \frac{
            \sum_{p \in P} 
            \sqrt{p_p \times \sqrt[|A_p|]{\prod_{k\in A_p}a_{p,k}}}
            }{|P|} \\
        \text{otherwise}: & 1
    \end{array}
\right. \\
\end{multline*}
This is repeated for all queries and all detected boxes.

The above scoring function combines the scores of the object, object-attribute, parts and part-attributes mentioned in the query. Note that the first part of the scoring function only combines object and object-attribute scores, while the second part combines part and part-attribute scores. While combining part scores we use a combination of arithmetic and geometric means. We found this combination to provide the best results empirically.

\begin{table}
    \setlength{\tabcolsep}{2mm}  
    \centering
    \footnotesize
    \begin{tabular}{c||c |c }
    \hline
    & mask $AP$ & box $AP$ \\
    Model & \apobj & \apobj \\
    \hline \hline
    R50 FPN & 31.2 \deemph{$\pm$ 0.1} & 34.3 \deemph{$\pm$ 0.2} \\
    \hline
    R50 FPN - object only  & 32.4 \deemph{$\pm$ 0.6} & 35.5 \deemph{$\pm$ 0.5} \\
    \hline
    ViT-L FPN & 44.7 \deemph{$\pm$ 0.4} & 49.6 \deemph{$\pm$ 0.4} \\
    \hline
    ViT-L FPN - object only & 39.8 \deemph{$\pm$ 0.1} & 43.6 \deemph{$\pm$ 0.1} \\
    \hline
    \end{tabular}
    \caption{Comparison of model performance on object segmentation when trained only on the task of object segmentation vs joint training on object and part segmentation and attribute prediction.}
    \label{tab:obj_only}
\end{table}

\begin{table*}
    \centering
    \footnotesize
    \begin{tabular}{c || c c || c c || c c || c c }
  & \multicolumn{2}{c||}{L1 queries} & \multicolumn{2}{c||}{L2 queries} & \multicolumn{2}{c||}{L3 queries} & \multicolumn{2}{c}{all queries}\\
    \hline
    Model & $AR@1$ & $AR@5$ & $AR@1$ & $AR@5$ & $AR@1$ & $AR@5$ & $AR@1$ & $AR@5$ \\
    \hline \hline
    R50 FPN & 22.5 \deemph{$\pm$ 0.7} & 39.2 \deemph{$\pm$ 0.5} & 20.1 \deemph{$\pm$ 0.4} & 38.5 \deemph{$\pm$ 0.1} & 22.3 \deemph{$\pm$ 0.9} & 44.5 \deemph{$\pm$ 1.1} & 21.4 \deemph{$\pm$ 0.6} & 40.9 \deemph{$\pm$ 0.3} \\
    + cascade & 23.5 \deemph{$\pm$ 1.4} & 41.1 \deemph{$\pm$ 2.7} & 21.4 \deemph{$\pm$ 2.4} & 40.9 \deemph{$\pm$ 3.2} & 25.3 \deemph{$\pm$ 2.7} & 48.1 \deemph{$\pm$ 3.2} & 23.3 \deemph{$\pm$ 2.3} & 43.7 \deemph{$\pm$ 3.1} \\
    \hline
    R101 FPN & 23.1 \deemph{$\pm$ 0.7} & 40.5 \deemph{$\pm$ 1.4} & 20.0 \deemph{$\pm$ 0.6} & 39.3 \deemph{$\pm$ 1.0} & 23.1 \deemph{$\pm$ 0.7} & 45.2 \deemph{$\pm$ 0.6} & 21.7 \deemph{$\pm$ 0.6} & 41.8 \deemph{$\pm$ 0.8} \\
    + cascade & 26.3 \deemph{$\pm$ 0.4} & 45.1 \deemph{$\pm$ 0.5} & 24.0 \deemph{$\pm$ 0.1} & 43.2 \deemph{$\pm$ 0.1} & 26.6 \deemph{$\pm$ 1.2} & 49.5 \deemph{$\pm$ 0.8} & 25.4 \deemph{$\pm$ 0.5} & 45.9 \deemph{$\pm$ 0.4} \\
    \hline
    ViT-B FPN & 26.8 \deemph{$\pm$ 0.2} & 45.8 \deemph{$\pm$ 0.2} & 22.7 \deemph{$\pm$ 0.5} & 40.0 \deemph{$\pm$ 0.7} & 24.1 \deemph{$\pm$ 0.5} & 42.5 \deemph{$\pm$ 1.5} & 23.9 \deemph{$\pm$ 0.4} & 42.0 \deemph{$\pm$ 0.9} \\
    + cascade & 27.0 \deemph{$\pm$ 0.4} & 46.1 \deemph{$\pm$ 0.5} & 23.0 \deemph{$\pm$ 0.9} & 40.3 \deemph{$\pm$ 0.2} & 25.5 \deemph{$\pm$ 0.8} & 43.1 \deemph{$\pm$ 0.5} & 24.7 \deemph{$\pm$ 0.7} & 42.4 \deemph{$\pm$ 0.2} \\
    \hline
    ViT-L FPN & 35.3 \deemph{$\pm$ 0.7} & 57.3 \deemph{$\pm$ 0.6} & 29.7 \deemph{$\pm$ 0.6} & 50.1 \deemph{$\pm$ 0.2} & 31.1 \deemph{$\pm$ 0.8} & 52.3 \deemph{$\pm$ 0.9} & 31.2 \deemph{$\pm$ 0.4} & 52.2 \deemph{$\pm$ 0.5} \\
    + cascade & 33.8 \deemph{$\pm$ 0.7} & 57.2 \deemph{$\pm$ 0.2} & 29.0 \deemph{$\pm$ 0.7} & 50.2 \deemph{$\pm$ 0.2} & 30.1 \deemph{$\pm$ 0.7} & 51.8 \deemph{$\pm$ 1.8} & 30.2 \deemph{$\pm$ 0.6} & 52.0 \deemph{$\pm$ 0.6} \\
    \end{tabular}
    \caption{Zero-shot instance detection results for different query levels for FPN and cascade models from Sec.~\attrexp{} trained and evaluated on \datalvis{}.}
    \label{tab:zero_shot_inst_det_lvis}
\end{table*}

\begin{table*}
    \centering
    \footnotesize
    \begin{tabular}{c c || c c || c c || c c || c c }
  & & \multicolumn{2}{c||}{L1 queries} & \multicolumn{2}{c||}{L2 queries} & \multicolumn{2}{c||}{L3 queries} & \multicolumn{2}{c}{all queries}\\
    \hline
    Model & Eval set & $AR@1$ & $AR@5$ & $AR@1$ & $AR@5$ & $AR@1$ & $AR@5$ & $AR@1$ & $AR@5$ \\
    \hline \hline
    R50 FPN & \datalvis{} & 22.0 \deemph{$\pm$ 0.4} & 39.6 \deemph{$\pm$ 0.6} & 20.6 \deemph{$\pm$ 0.5} & 39.0 \deemph{$\pm$ 0.7} & 24.7 \deemph{$\pm$ 1.0} & 45.5 \deemph{$\pm$ 1.4} & 22.4 \deemph{$\pm$ 0.3} & 41.6 \deemph{$\pm$ 0.7} \\
    R101 FPN & \datalvis{} & 23.5 \deemph{$\pm$ 0.5} & 40.9 \deemph{$\pm$ 0.4} & 21.2 \deemph{$\pm$ 0.3} & 40.1 \deemph{$\pm$ 0.7} & 24.3 \deemph{$\pm$ 1.3} & 45.2 \deemph{$\pm$ 0.9} & 22.8 \deemph{$\pm$ 0.5} & 42.2 \deemph{$\pm$ 0.6} \\
    ViT-B FPN & \datalvis{} & 29.5 \deemph{$\pm$ 0.6} & 49.5 \deemph{$\pm$ 1.1} & 25.8 \deemph{$\pm$ 1.4} & 44.9 \deemph{$\pm$ 2.3} & 26.2 \deemph{$\pm$ 1.2} & 45.7 \deemph{$\pm$ 2.9} & 26.6 \deemph{$\pm$ 1.1} & 46.0 \deemph{$\pm$ 2.2} \\
    ViT-L FPN & \datalvis{} & 38.0 \deemph{$\pm$ 0.6} & 60.8 \deemph{$\pm$ 1.2} & 33.3 \deemph{$\pm$ 1.7} & 55.6 \deemph{$\pm$ 1.9} & 33.1 \deemph{$\pm$ 2.6} & 59.0 \deemph{$\pm$ 2.8} & 34.0 \deemph{$\pm$ 1.8} & 57.8 \deemph{$\pm$ 2.1} \\
    \hline
    R50 FPN & \dataego{} & 15.4 \deemph{$\pm$ 0.1} & 29.1 \deemph{$\pm$ 0.6} & 13.2 \deemph{$\pm$ 0.2} & 28.0 \deemph{$\pm$ 0.9} & 14.4 \deemph{$\pm$ 1.8} & 29.1 \deemph{$\pm$ 1.3} & 14.2 \deemph{$\pm$ 0.9} & 28.7 \deemph{$\pm$ 0.8} \\
    R101 FPN & \dataego{} & 16.3 \deemph{$\pm$ 0.5} & 29.8 \deemph{$\pm$ 0.9} & 15.0 \deemph{$\pm$ 0.6} & 28.6 \deemph{$\pm$ 0.7} & 14.2 \deemph{$\pm$ 0.6} & 28.3 \deemph{$\pm$ 0.9} & 14.9 \deemph{$\pm$ 0.1} & 28.6 \deemph{$\pm$ 0.5} \\
    ViT-B FPN & \dataego{} & 13.5 \deemph{$\pm$ 1.2} & 24.4 \deemph{$\pm$ 1.3} & 11.0 \deemph{$\pm$ 0.4} & 19.5 \deemph{$\pm$ 0.7} & 9.3 \deemph{$\pm$ 0.5} & 18.1 \deemph{$\pm$ 0.5} & 10.6 \deemph{$\pm$ 0.1} & 19.7 \deemph{$\pm$ 0.4} \\
    ViT-L FPN & \dataego{} & 20.8 \deemph{$\pm$ 0.2} & 36.9 \deemph{$\pm$ 0.7} & 19.8 \deemph{$\pm$ 1.3} & 33.3 \deemph{$\pm$ 1.3} & 21.4 \deemph{$\pm$ 1.2} & 34.9 \deemph{$\pm$ 0.7} & 20.7 \deemph{$\pm$ 1.0} & 34.7 \deemph{$\pm$ 0.9} \\
    \end{tabular}
    \caption{Zero-shot instance detection results for different query levels for FPN models from Sec.~\attrexp{} trained on joint \dataname{} dataset and evaluated on \datalvis{} and \dataego{}.}
    \label{tab:zero_shot_inst_det_joint}
\end{table*}

\begin{table}
    \centering
    \footnotesize
    \begin{tabular}{c || c c }
    & \multicolumn{2}{c}{all queries}\\
    \hline
    Score components & $AR@1$ & $AR@5$ \\
    \hline \hline
    Object only & 1.9 \deemph{$\pm$ 0.5} & 8.2 \deemph{$\pm$ 0.2} \\
    \hline
    Object + part & 2.4 \deemph{$\pm$ 0.4} & 10.8 \deemph{$\pm$ 0.9} \\
    \hline
    Object + color & 5.6 \deemph{$\pm$ 0.5} & 15.5 \deemph{$\pm$ 0.1} \\
    \hline
    Object + attribute & 8.5 \deemph{$\pm$ 0.4} & 22.3 \deemph{$\pm$ 0.2} \\
    \hline
    Object + part + color & 20.8 \deemph{$\pm$ 0.6} & 40.2 \deemph{$\pm$ 0.6} \\
    \hline
    All & 31.2 \deemph{$\pm$ 0.4} & 52.2 \deemph{$\pm$ 0.5} \\
    \end{tabular}
    \vspace*{0.05in}
    \caption{Ablation study on importance of object, part, and attribute predictions on zero-shot instance detection performance. We compute metrics using only object, object + part, object + color, object + attribute, object + part + color, and all ViT-L FPN model scores.}
    \label{tab:zero_shot_ablation}
\end{table}

\section{Evaluation of open world detectors on zero-shot instance detection task}
\label{open_world_det_eval}
In this section we give details on how we evaluated Detic~\cite{zhou2022detecting} and MDETR~\cite{kamath2021mdetr} on zero-shot instance detection task. For both projects we used code open sourced on GitHub.

Detic supports a custom vocabulary and encodes natural language class descriptions using pre-trained CLIP text encoder. We used all $5k$ queries as custom vocabulary so that we have prediction scores for all queries for each detected box. Due to large vocabulary we had to increase the number of detections per image. We experimented with this parameter and found that $2,000$ boxes gives the best results. We used plain query strings (e.g., ``A dog with brown ear and black neck") from \dataname{} dataset as class descriptions along with 3 more prompt variants with prefixes ``A photo of", ``A close up picture of", and ``A close up photo of" in front of the plain query strings. The ``close up" variants were an attempt to guide text embeddings closer to a detection setup but we didn't see much improvement in performance. We use {\fontfamily{qcr}\selectfont Detic\_LCOCOI21k\_CLIP\_SwinB\_896b32\_4x\_ft4x\_max-size.pth} model and report mean and standard deviation $AR@k$ calculated over results from these $4$ prompt variants.


MDETR is geared towards referring expressions and phrase grounding and treats each image-text pair independently. We follow inference similar to LVIS evaluation reported in the MDETR paper. Namely, for inference on a given image, we evaluate the model on each of the $5k$ queries separately, then merge the sets of boxes detected on each of the queries and keep the boxes corresponding to top K query scores. Unlike Detic, predicted boxes are not shared across queries since MDETR predicts bounding boxes independently for each query. As a result, we had to increase the number of detections per image even further to $10,000$ to obtain the best results. We also experimented with two MDETR models with R101 backbone \footnote{A known issue (\#86) prevented the use of ENB backbones}, one trained for referring expressions task ({\fontfamily{qcr}\selectfont refcocog\_resnet101\_checkpoint.pth}) and the other for LVIS few-shot task ({\fontfamily{qcr}\selectfont lvis10\_checkpoint.pth}) and observed that LVIS few-shot task model performs better. We report mean and standard deviation of results from that model over the same $4$ query prompt variants as Detic. 

\section{Few-shot instance detection experiments}
\label{few_shot_exp}
\begin{figure}
    \centering
    \includegraphics[width=1.0\columnwidth]{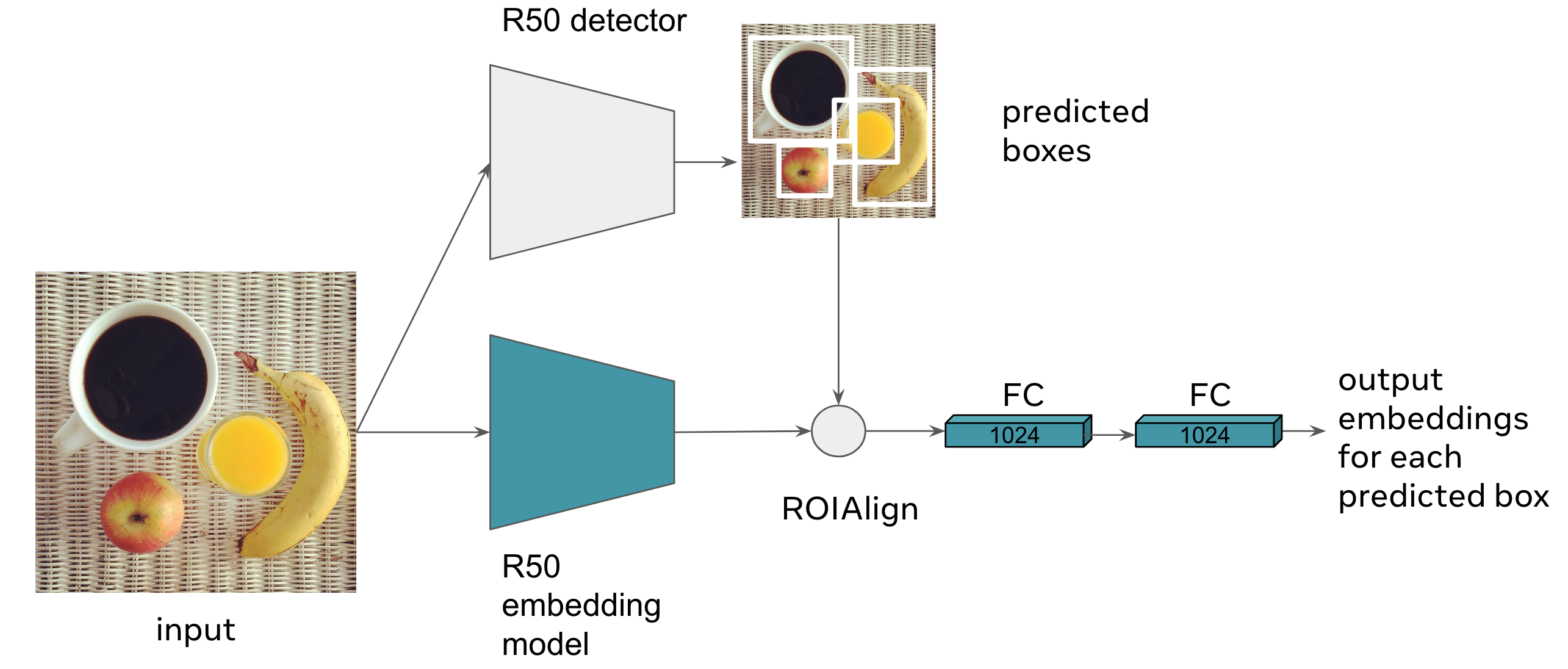}
    \caption{
    The few-shot instance detection model consists of a frozen detector and an embedding model. The detector outputs class-agnostic bounding boxes. The embedding model takes an image and a set of predicted bounding boxes on the image as inputs, and outputs embeddings for every box.}
    \label{fig:fewshot_inference}
\end{figure}

The few-shot model is a two-tower model as shown in Fig. ~\ref{fig:fewshot_inference}, where the (a) first tower is a detection model which predicts object boxes in the images and (b) the second tower is an embedding model that provides a feature embedding for each of the predicted boxes. The two towers are learned independently.

\noindent{\textbf{Query feature registration.}} In the few-shot setup, for each query $Q$ we are provided a set of ``query images" with one bounding box per image for the ``query" object instance. We first extract a feature for each of query boxes only using the embedding model. Given $k$ query images (with bounding box) for the query $Q$, we extract $k$ query features. The features are then averaged to obtain an average query feature vector $f_{Q}$.

\noindent{\textbf{Instance detection with query features.}}
We are also provided a set of target images for each query $\mathcal{I}_Q$ from which another bounding box corresponding to the query needs to be extracted. For each image $I \in \mathcal{I}_Q$, we first predict $100$ bounding boxes $\mathcal{B}_I$ using the detection tower of our model. Each of these boxes $B \in \mathcal{B}_I$ are then represented by a feature vector $f_{B,I}$ using the embedding model. All the boxes are then ranked based on the cosine similarity of their feature with the query feature $f_Q$. The top $N$ returned boxes from $\mathcal{I}_Q$ are used to compute AR@N for $N=1,5$.

\noindent{\textbf{Detection model.}} We train a standard R50-FPN mask R-CNN model with $75$ object categories on the \texttt{train} split of the PACO dataset. During the feature registration and instance detection stage, we ignore the category label and only use the predicted boxes.

\begin{figure}
    \centering
    \includegraphics[width=1.0\columnwidth]{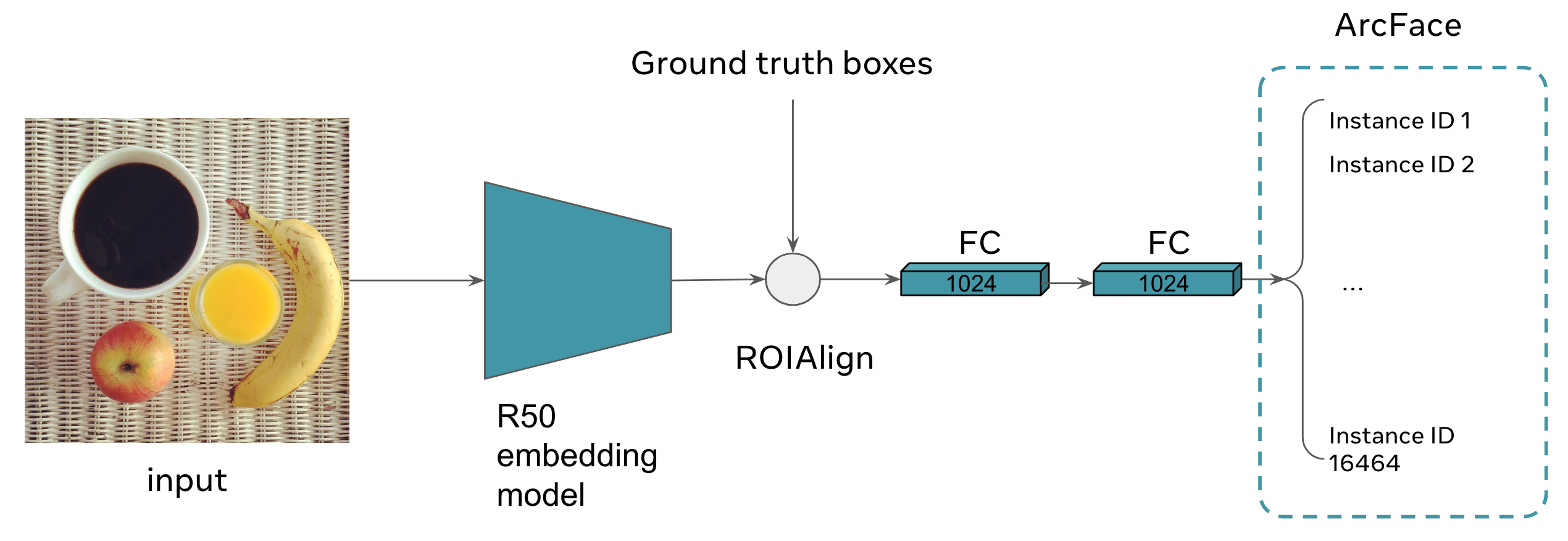}
    \caption{The embedding model is a mask R-CNN style model with a custom ROI head where the softmax loss is replaced with an ArcFace loss using instance IDs as supervision for richer representations for instance recognition. Once training is finished, we throw away the ArcFace layer and use the outputs from the last FC layer as per-box representations.}
    \label{fig:fewshot_train}
\end{figure}

\noindent{\textbf{Embedding model.}} The embedding model is a mask R-CNN style model with a custom ROI head as shown in Fig. ~\ref{fig:fewshot_train}. During inference, it takes predicted bounding boxes as input and outputs embeddings for each box with ROIAlign~\cite{he2017mask}. The model is trained with ArcFace~\cite{deng2019arcface} loss to have richer representations for instance recognition. We trained the embedding model with an ArcFace loss to perform 16464-way instance ID classification at box-level. The model was trained to distinguish the $16464$ different object instances in the PACO-Ego4d \texttt{train} split. In this dataset, each instance has multiple bounding boxes, making it possible to train such a model. We simply use ground truth boxes during training to avoid handling the additional complexity from distinguishing foreground and background boxes. Note that the sets of instances in \texttt{train} and \texttt{test} splits are completely disjoint.


{\noindent{\textbf{Implementation details.}}}
We use R50-FPN~\cite{lin2017feature} as the backbone. The custom ROI head is implemented as a ROIAlign operator followed by 2 FC layers with 1024 dimensions. The ArcFace layer is configured with \emph{margin = 0.5} and \emph{scale = 8.0}. We use the default data augmentation for Faster R-CNN~\cite{ren2015faster} training in Detectron2~\cite{wu2019detectron2}. We train the embedding model on the PACO-Ego4D \texttt{train} split for 22.5K iterations. We set \emph{lr = 0.04} and use Cosine \emph{lr} decay. The batch size is 128 distributed across 32 GPUs (4 images per GPU).

